\pdfoutput=1
\documentclass[conference]{IEEEtran}
\usepackage[utf8]{inputenc} 
\usepackage[T1]{fontenc}    
\usepackage{booktabs}       
\usepackage{amsfonts}       
\usepackage{nicefrac}       
\usepackage{microtype}      
\usepackage{amsmath}

\let\labelindent\relax
\usepackage{enumitem}

\usepackage{times}
\usepackage[numbers]{natbib}
\usepackage[bookmarks=true]{hyperref}
\usepackage[acronym]{glossaries}
\usepackage[pdftex]{graphicx}
\usepackage{subcaption}
\usepackage{url}
\usepackage{tabularx}
\usepackage{algorithm}
\usepackage[noend]{algpseudocode}
\usepackage{mathtools}
\usepackage{multirow}
\usepackage{multicol}
\usepackage{xcolor}
\usepackage[colorinlistoftodos]{todonotes}

\usepackage{amsmath,amsfonts,bm}

\def\Tabref#1{Table~\ref{#1}}

\def\Figref#1{Figure~\ref{#1}}

\def\Secref#1{Section~\ref{#1}}

\def\eqref#1{equation~\ref{#1}}
\def\Eqref#1{Equation~\ref{#1}}

\def\1{\bm{1}}

\def\eps{{\epsilon}}

\def\rz{{\textnormal{z}}}

\def\rva{{\mathbf{a}}}

\def\rvs{{\mathbf{s}}}

\def\vtheta{{\bm{\theta}}}
\def\va{{\bm{a}}}

\def\vs{{\bm{s}}}

\DeclareMathAlphabet{\mathsfit}{\encodingdefault}{\sfdefault}{m}{sl}
\SetMathAlphabet{\mathsfit}{bold}{\encodingdefault}{\sfdefault}{bx}{n}

\newcommand{\E}{\mathbb{E}}

\newcommand{\R}{\mathbb{R}}

\graphicspath{{figures/}}

\newcommand{\vphi}{\bm{\phi}}
\newcommand{\vpsi}{\bm{\psi}}
\newcommand{\target}[1]{\bar{#1}}

\newcommand{\uniform}[2]{\mathcal{U}\left({#1},{#2}\right)}
\newcommand{\normal}[2]{\mathcal{N}\left({#1},{#2}\right)}
\newcommand{\lognormal}[2]{\mathrm{Lognormal}\left({#1},{#2}\right)}
\newcommand{\bernouilli}[1]{\mathcal{B}\left({#1}\right)}
\newcommand{\expnumber}[2]{{#1}\mathrm{e}{#2}}
\newcommand{\bE}{\mathbb{E}}

\newacronym{rl}{RL}{Reinforcement Learning}
\newacronym{morl}{MO\gls{rl}}{Multi-Objective \gls{rl}}
\newacronym[\glslongpluralkey={Markov Decision Processes}]{mdp}{MDP}{Markov Decision Process}
\newacronym[\glslongpluralkey={Constrained Markov Decision Processes}]{cmdp}{CMDP}{Constrained Markov Decision Process}
\newacronym{mpo}{MPO}{Maximum a Posteriori Policy Optimisation}
\newacronym{td}{TD}{Temporal Difference}
\newacronym[\glslongpluralkey={Degrees of Freedom}]{dof}{DoF}{Degree of Freedom}

\newcommand{\red}[1]{\textcolor{red}{\textbf{#1}}}
\newcommand{\green}[1]{\textcolor{green}{\textbf{#1}}}

\begin{document}

\title{Value constrained model-free continuous control}

\author{
\authorblockN{Steven Bohez, Abbas Abdolmaleki, Michael Neunert, Jonas Buchli, Nicolas Heess \& Raia Hadsell}
\authorblockA{DeepMind, London, UK}
}

\maketitle

\begin{abstract}
The naive application of \acrlong{rl} algorithms to continuous control problems -- such as locomotion and manipulation -- often results in policies which rely on high-amplitude, high-frequency control signals, known colloquially as \emph{bang-bang} control.
Although such solutions may indeed maximize task reward, they can be unsuitable for real world systems.
Bang-bang control may lead to increased wear and tear or energy consumption, and tends to excite undesired second-order dynamics. 
To counteract this issue, multi-objective optimization can be used to simultaneously optimize both the reward and some auxiliary cost that discourages undesired (e.g. high-amplitude) control.
In principle, such an approach can yield the sought after, smooth, control policies.
It can, however, be hard to find the correct trade-off between cost and return that results in the desired behavior.
In this paper we propose a new constraint-based reinforcement learning approach that ensures task success while minimizing one or more auxiliary costs (such as control effort).
We employ Lagrangian relaxation to learn both (a) the parameters of a control policy that satisfies the desired constraints and (b) the Lagrangian multipliers for the optimization.
Moreover, we demonstrate that we can satisfy constraints either in expectation or in a per-step fashion, and can even learn a single policy that is able to dynamically trade-off between return and cost.
We demonstrate the efficacy of our approach using a number of continuous control benchmark tasks, a realistic, energy-optimized quadruped locomotion task, as well as a reaching task on a real robot arm.\footnote{Videos available at \url{https://sites.google.com/view/successatanycost}}
\end{abstract}

\section{Introduction}

Deep \gls{rl} has achieved great successes over the last years, enabling learning of effective policies from high-dimensional input, such as pixels, on complicated tasks.
However, compared to problems with discrete action spaces, continuous control problems with high-dimensional continuous state-action spaces -- such as those encountered in robotics -- have proven much more challenging.
One problem encountered in continuous action spaces is that straightforward optimization of task reward leads to idiosyncratic solutions that switch between extreme values for the controls at high-frequency, a phenomenon also referred to as \emph{bang-bang} control.
While such solutions can maximize reward and can be acceptable in simulation, they are usually not suitable for real-world systems where smooth control signals are desirable.
Unnecessary oscillations are not only energy inefficient, they also exert stress on a physical system by exciting second-order dynamics and increasing wear and tear.

To regularize the behavior, a common strategy is to add penalties to the reward function.
As a result, the reward function is composed of positive reward for achieving the goal and negative reward (cost) for high control actions or high energy usage.
This effectively casts the problem into a multi-objective optimization setting, where -- depending on the ratio between the reward and the different penalties -- different behaviors may emerge.
While every ratio will have its optimal policy, finding the ratio that results in the desired behavior can be difficult.
Often, one must find different hyperparameter settings for different reward-penalty trade-offs or tasks.
The process of finding these parameters can be cumbersome, and may prevent robust and general solutions. 
In this paper we rephrase the problem: instead of trying to find the appropriate ratio between reward and cost, we regularize the optimization problem by adding constraints, thereby reducing its effective dimensionality.
More specifically, we propose to minimize the penalty while respecting a lower-bound on the success rate of the task.

Using a Lagrangian relaxation technique, we introduce cost coefficients for each of the imposed constraints that are tuned automatically during the optimization process.
In this way we can find the optimal trade-off between reward and costs (that also satisfies the imposed constraints) automatically.
By making the cost multipliers state-dependent, and adapting them alongside the policy, we can not only impose constraints on expected reward or cost, but also on their instantaneous values.
Such point-wise constraints allow for much tighter control over the behavior of the policy, since a constraint that is satisfied only in overall expectation could still be violated momentarily.
Our constrained optimization procedure can further be generalized to a multitask setting to train policies that are able to dynamically trade-off reward and penalties within and across tasks.
This allows us to, for example, learn energy-efficient locomotion at a range of different velocities.

The contributions of this work are (i) we demonstrate that state-dependent Lagrangian multipliers for large and continuous state spaces can be implemented with a neural network that generalizes across states; (ii) we introduce a structured critic that simultaneously learns both reward and value estimates as well as the coefficient to balance them in a single model; and finally (iii) demonstrate how our constrained optimization framework can be employed in a multi-task setting to effectively train goal-conditioned policies.
Our approach is general and flexible in that it can be applied to any value-based \gls{rl} algorithm and any number of constraints.
We evaluate our approach on a number of simulated continuous control problems in \Secref{sec:experiments} using tasks from the DM Control Suite~\citep{Tassa2018ControlSuite} and a (realistically simulated) locomotion task with the Minitaur quadruped.
Finally, we apply our method to a reaching task that requires satisfying a visually defined constraint on a real Sawyer robot arm.

\section{Background and related work}

We consider \glspl{mdp} \citep{Sutton1998RL} where an agent sequentially interacts with an environment, in each step observing the state of the environment $\vs$ and choosing an action according to a policy $\va \sim \pi\left(\vs \mid \vs\right)$.
Executing the action in the environment causes a state transition with an associated reward defined by some utility function $r\left(\vs,\va\right)$.
The goal of the agent is to maximize the expected sum of rewards along trajectories generated by following policy $\pi$, also known as the return, $\max_{\pi} \E_{\rvs,\rva \sim \pi}\left[\sum_{t} r\left(\vs_{t},\va_{t}\right)\right]$.
While some tasks have well-defined reward (e.g. the increase in score when playing a game) in other cases it is up to the user to define a reward function that produces a desired behavior.
Designing suitable reward functions can be a difficult problem even in the single-objective case \citep[e.g.][]{popov2017data,amodei2016concrete}.

\gls{morl} problems arise in many domains, including robotics, and have been covered by a rich body of literature \citep[see e.g.][for a recent review]{Roijers2013_MORL}, suggesting a variety of solution strategies.
For instance, \cite{MossalamARW16} devise a Deep \gls{rl} algorithm that implements an outer loop method and repeatedly calls a single-objective solver.
\cite{MannorJMLR2004} propose an algorithm for learning in a stochastic game setting with vector valued rewards (their approach is based on approachability of a target set in the reward space).
However, most of these approaches explicitly recast the multi-objective problem into a single-objective problem (that is amenable to existing methods), where one aims to find the trade-off between the different objectives that yields the desired result.
In contrast, we aim for a method that automatically trades off different components in the objective to achieve a particular goal.
To achieve this, we cast the problem in the framework of \glspl{cmdp} \citep{altman1999constrained}.
\glspl{cmdp} have been considered in a variety of works, including in the robotics and control literature.
For instance, \citet{Achiam2017_CPO} and \citet{dalal2018safe} focus on constraints motivated by safety concerns and propose algorithms that ensure that constraints remain satisfied at all times.
These works, however, assume that the initial policy already satisfies the constraint, which is usually not practical when, as in our case, the constraint involves the task success rate.
The motivation for the work by \cite{Tessler2018_RCPO} is similar to ours, but unlike ours their approach maximizes reward subject to a constraint on the cost and enforces constraints only in expectation.

Constraint-based formulations are also frequently used in single-objective policy search algorithms where bounds on the policy divergence are employed to control the rate of change in the policy from one iteration to the next \cite[e.g.][]{peters2010relative,Levine2013GPS,schulman15trpo,abdolmaleki2018maximum}.
Our use of constraints, although similar in the practical implementation, is conceptually orthogonal.
Also these methods typically employ constraints that are satisified only in expectation. 
While we note that our approach can be applied to any value-based off-policy method, we make use of the method described in \gls{mpo}  \citep{abdolmaleki2018maximum} as the underlying policy optimization algorithm -- without loss of any generality of our method.
\gls{mpo} is an actor-critic algorithm that is known to yield robust policy improvement.
In each policy improvement step, for each state sampled from replay buffer, \gls{mpo} creates a population of actions.
Subsequently, these actions are re-weighted based on their estimated values such that better actions will have higher weights.
Finally, \gls{mpo} uses a supervised learning step to fit a new policy in continuous state and action space.
See \citet{abdolmaleki2018maximum} and Appendix A for more details.

\section{Constrained optimization for control}
\label{sec:approach}

We consider \glspl{mdp} where we have both a reward and cost, $r\left(\vs,\va\right)$ and $c\left(\vs,\va\right)$, which are functions of state $\vs$ and action $\va$.
The goal is to automatically find a stochastic policy $\pi\left(\va | \vs ; \theta\right)$ (with parameter $\theta$) that both maximizes the (expected) reward that defines task success and minimizes a cost that regularizes the solution.
For instance, in the case of the well-known cart-pole problem we might want to achieve a stable swing-up while minimizing other quantities, such as control effort or energy.
This can be expressed using a penalty proportional to the total cost, i.e. $
    \max_{\pi} \E_{\rvs,\rva \sim \pi}\left[\sum_{t} r\left(\vs_{t},\va_{t}\right) - \alpha \cdot c\left(\vs_{t},\va_{t}\right)\right]
$, where we take $\max_\pi$ to mean maximizing the objective with respect to the policy parameters $\theta$ and the expectation is with respect to trajectories produced by executing policy $\pi$.
The problem then becomes one of finding an appropriate trade-off between task reward and cost, and hence a suitable value of $\alpha$.
Finding this trade-off is often non-trivial.
An alternative way of looking at this dilemma is to take a multi-objective optimization perspective.
Instead of fixing $\alpha$, we can optimize for it simultaneously and can obtain different Pareto-optimal solutions for different values of $\alpha$.
In addition, to ease the definition of a desirable regime for $\alpha$, one can consider imposing hard constraints on the cost to reduce dimensionality \citep{deb2014multi}, instead of linearly combining the different objectives.
Defining such hard constraints is often more intuitive than trying to manually tune coefficients.
For example, in locomotion, it is easier to define desired behavior in terms of a lower bound on speed or an upper bound on an energy cost.

\subsection{Constrained MDPs}
\label{sec:cmdp}

The perspective outlined above can be formalized as a \gls{cmdp}~\citep{altman1999constrained}.
A constraint can be placed on either the reward or the cost.
In this work we primarily consider a lower bound on the expected total return (although the theory derived below equivalently applies to constraints on cost, i.e. $
    \min_{\pi} \E_{\rvs,\rva \sim \pi}\left[\sum_{t}c\left(\vs_{t},\va_{t}\right)\right] \text{, s.t. } \E_{\rvs,\rva \sim \pi}\left[\sum_{t} r\left(\vs_{t},\va_{t}\right)\right] \geq \target{R}
$, where $\target{R}$ is the minimum desired return.
In the case of an infinite horizon with a given stationary state distribution, the constraint can instead be formulated for the per-step reward, i.e. $\E_{\rvs,\rva \sim \pi}\left[r\left(\vs,\va\right)\right] \geq \target{r}$.
In practice one often optimizes the $\gamma$-discounted return in both cases.
To apply model-free \gls{rl} methods to this problem we first define an estimate of the expected discounted return for a given policy as the action-value function $Q_{r}\left(\vs,\va\right) = \E_{\rvs,\rva \sim \pi}\left[\sum_{t} \gamma^t \cdot r\left(\vs_{t},\va_{t}\right)\vert \vs_{0} = \vs, \va_{0} = \va\right]$, and similarly the expected discounted cost $Q_{c}\left(\vs,\va\right)$.
We can then recast the \gls{cmdp} in value-space, where $\target{V_{r}} = \target{r}/\left(1-\gamma\right)$ (i.e. scaling the desired reward $\target{r}$ with the limit of the converging sum over discounts):
\begin{equation}
    \label{eq:cmdp}
    \min_{\pi} \E_{\rvs,\rva \sim \pi}\left[Q_{c}\left(\vs,\va\right)\right]\text{, s.t. } \E_{\rvs,\rva \sim \pi}\left[Q_{r}\left(\vs,\va\right)\right] \geq \target{V_{r}}\text{.}
\end{equation}

\subsection{Lagrangian relaxation}
\label{sec:lag_rel}

We formulate task success as a lower bound on the reward.
This constraint is typically not satisfied at the start of learning since the agent first needs to learn how to solve the task.
This rules out methods for solving \glspl{cmdp} which assume that the constraint is satisfied at the start and limit the policy update to remain within the constraint-satisfying regime \citep[e.g.][]{Achiam2017_CPO}.
Lagrangian relaxation is a general method for solving general constrained optimization problems; and \glspl{cmdp} by extension \citep{altman1999constrained}.
In this setting, the hard constraint is relaxed into a soft constraint, where any constraint violation acts as a penalty for the optimization.
Applying Lagrangian relaxation to \Eqref{eq:cmdp} results in the unconstrained dual problem 
\begin{multline}
\label{eq:relaxation}
    \max_{\pi} \min_{\lambda \geq 0} \E_{\rvs,\rva \sim \pi}\left[Q_{\lambda}\left(\vs,\va\right)\right] \text{,}\\
    \text{with } Q_{\lambda}\left(\vs,\va\right) = \lambda\left(Q_{r}\left(\vs,\va\right)-\target{V_{r}}\right)-Q_{c}\left(\vs,\va\right)\text{,}
\end{multline}
with an additional minimization w.r.t.~the multiplier $\lambda$.

A larger $\lambda$ results in a higher penalty for violating the constraint.
Hence, we can iteratively update $\lambda$ by gradient descent on $Q_{\lambda}\left(\vs,\va\right)$, and alternate with policy optimization, until the constraint is satisfied.
Under assumptions described in \citet{Tessler2018_RCPO}, this approach converges to a saddle point.
At convergence, when $\nabla_{\lambda}\E\left[Q_{\lambda}\left(\vs,\va\right)\right] = 0$, $\lambda$ is exactly the desired trade-off between reward and cost we aimed to find.
To perform the policy optimization for $\pi$ any off-the-shelf off-policy optimization algorithm can be used (since we assume that we have a learned, approximate Q-function at our disposal).
In practice, we perform policy optimization using the MPO algorithm \citep{abdolmaleki2018maximum} and refer to Appendix A for additional details.

At the start of learning, as the constraint is not yet satisfied, $\lambda$ will grow in order to suppress the cost $Q_{c}\left(\vs,\va\right)$ and focus the optimization on maximizing $Q_{r}\left(\vs,\va\right)$.
Depending on how quickly the constraint can be satisfied, $\lambda$ can grow very large, resulting in a overall large magnitude of $Q_{\lambda}\left(\vs,\va\right)$.
This can result in unstable learning as most actor-critic methods that have an explicit parameterization of $\pi$ are especially sensitive to large (swings in) values.
To improve stability, we re-parameterize $Q_{\lambda}\left(\vs,\va\right)$ to be a projection into a convex combination of $\left(Q_{r}\left(\vs,\va\right)-\target{V_{r}}\right)$ and $-Q_{c}\left(\vs,\va\right)$.
Instead of scaling only the reward term, we can then adaptively reweight the relative importance of reward and cost, and make the magnitude of $Q_{\lambda}\left(\vs,\va\right)$ bounded.
To enforce $\lambda \geq 0$, we can perform a change of variable $\lambda' = \log\left(\lambda\right)$ to obtain the following dual optimization problem
\begin{multline}
\label{eq:normalisation}
    \max_{\pi} \min_{\lambda' \in \R} \E_{\rvs,\rva \sim \pi}\left[Q_{\lambda'}\left(\vs,\va\right)\right] \text{,}\\
    \text{with } Q_{\lambda'}\left(\vs,\va\right) = \frac{ \exp\left(\lambda'\right)\left(Q_{r}\left(\vs,\va\right)-V^*_{r}\right)-Q_{c}\left(\vs,\va\right)}{\exp\left(\lambda'\right) + 1}\text{.}
\end{multline}

Note that to correspond to the formulation in \Eqref{eq:relaxation}, we only perform gradient descent w.r.t. $\lambda'$ on the first term in the numerator.
In practice, we limit $\lambda'$ to $\left[\lambda_{\min}', \lambda_{\max}'\right]$, with $\left(\exp\left(\lambda_{\max}'\right) + 1\right)^{-1} = \eps$ for some small $\eps $, and initialize to $\lambda_{\max}'$.

\subsection{Point-wise constraints}

One downside of the \gls{cmdp} formulation given in \Eqref{eq:cmdp} is that the constraint is placed on the \emph{expected} total episode return, or \emph{expected} reward.
The constraint will therefore not necessarily be satisfied at every single timestep, or visited state, during the episode.
For some tasks this difference, however, turns out to be of importance.
For example, in locomotion, a constant speed is more desirable than a fluctuating one, even though the latter might also satisfy a minimum velocity in expectation.
Fortunately, we can extend the single constraint introduced in \Secref{sec:cmdp} to a set, possibly infinite, of point-wise constraints; one for each state induced by the policy.
This can be formulated as the following optimization problem:
\begin{equation}
    \label{eq:ext_cmdp}
    \min_{\pi} \E_{\rvs,\rva \sim \pi}\left[Q_{c}\left(\vs,\va\right)\right]\text{, s.t. } \forall \rvs \sim \pi : \E_{\rva \sim \pi}\left[Q_{r}\left(\vs,\va\right)\right] \geq \target{V_{r}}\text{.}
\end{equation}

Analogous to \Secref{sec:lag_rel}, this problem can be optimized with Lagrangian relaxation by introducing state-dependent Lagrangian multipliers.
Formally, we can write,
\begin{multline}
\label{eq:constraint_loss}
    \max_{\pi} \E_{\rvs \sim \pi}\left[\min_{\lambda\left(\vs\right) \geq 0} \E_{\rva \sim \pi}\left[Q_{\lambda}\left(\vs,\va\right)\right]\right] \text{,}\\
    \text{with } Q_{\lambda}\left(\vs,\va\right) = \lambda\left(\vs\right)\left(Q_{r}\left(\vs,\va\right)-\target{V_{r}}\right)-Q_{c}\left(\vs,\va\right)\text{.}
\end{multline}

Analogous to the assumption that nearby states have a similar value, here we assume that nearby states have similar $\lambda$ multipliers.
This allows learning a parametric function $\lambda\left(\vs\right)$ alongside the action-value which can generalize to unseen states $\vs$.
In practice, we train a single critic model that outputs $\lambda\left(\vs\right)$ as well as $Q_{c}\left(\vs,\va\right)$ and $Q_{r}\left(\vs,\va\right)$.
We provide pseudocode for the resulting constrained optimization algorithm in Appendix A.
Note that, in this case, the lower bound is still a fixed value and does not depend on the state.
In general such a constraint might be impossible to satisfy for some states in a given task if the state distribution is not stationary (e.g. we cannot satisfy a reward constraint in the swing-up phase of the simple pendulum).
However, the lower bound can also be made state-dependent and our approach will still be applicable.

\subsection{Conditional constraints}

Up to this point, we have made the assumption that we are only interested in a single, fixed value for the lower bound.
However, in some tasks one would want to solve \Eqref{eq:ext_cmdp} for different lower bounds $\target{V_{r}}$, i.e. minimizing cost for various success rates.
For example, in a locomotion task, one could be interested in optimizing energy for multiple different target speeds or gaits.
Assuming locomotion is a stationary behavior, one could set $\target{V_{r}} = \target{v}/\left(1-\gamma\right)$ for a range of velocities $\target{v} \in \left[0, \target{v_{\max}}\right]$.
In the limit this would achieve the same result as multi-objective optimization -- it would identify the set of solutions wherein it is impossible to increase one objective without worsening another -- also known as a Pareto front.
To avoid the need to solve a large number of optimization problems, i.e., solving for every $\target{V_{r}}$ separately, we can condition the policy, value function and Lagrangian multipliers on the desired target value and, effectively, learn a bound-conditioned policy
\begin{multline}
    \E_{\rz \sim p\left(\rz\right)} \left[ \max_{\pi\left(z\right)} \E_{\rvs \sim \pi\left(z\right)}\left[\min_{\lambda\left(\vs,z\right) \geq 0} \E_{\rva \sim \pi\left(z\right)}\left[Q_{\lambda}\left(\vs,\va,z\right)\right]\right]\right] \text{,}\\
    \text{with } Q_{\lambda}\left(\vs,\va,z\right) = \\
    \lambda\left(\vs,z\right) \left(Q_{r}\left(\vs,\va,z\right)-\target{V_{r}}\left(z\right)\right) -Q_{c}\left(\vs,\va,z\right)\text{.}
\end{multline}

Here $z$ is a goal variable, the desired lower bound for the reward, that is observed by the policy and critic and maps to a lower bound for the value $\target{V_{r}}\left(z\right)$.
Such a conditional constraint allows a single policy to dynamically trade off cost and return.

\begin{figure}[t!]
\centering
\begin{subfigure}[t]{0.24\columnwidth}
\includegraphics[width=\textwidth]{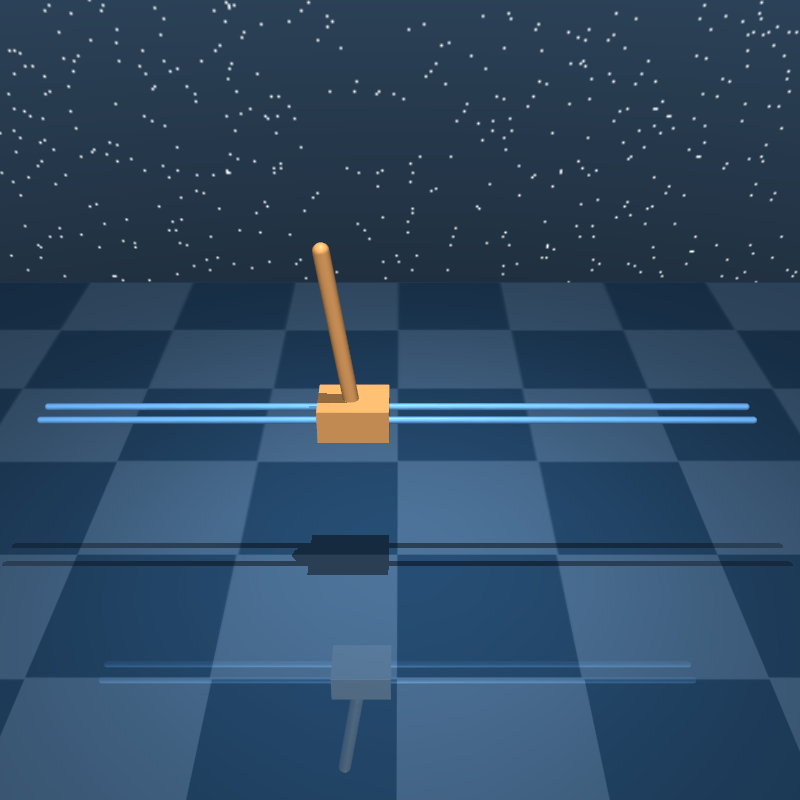}
\caption{Cart-pole}
\label{fig:cartpole_sim}
\end{subfigure}
\begin{subfigure}[t]{0.24\columnwidth}
\includegraphics[width=\textwidth]{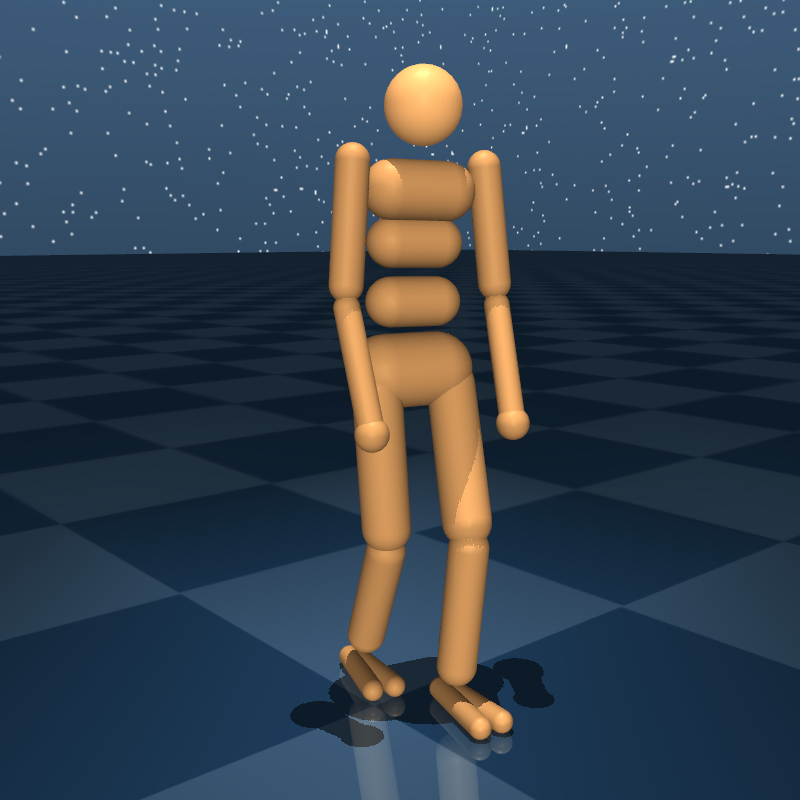}
\caption{Humanoid}
\label{fig:humanoid_sim}
\end{subfigure}
\begin{subfigure}[t]{0.24\columnwidth}
\includegraphics[width=\textwidth]{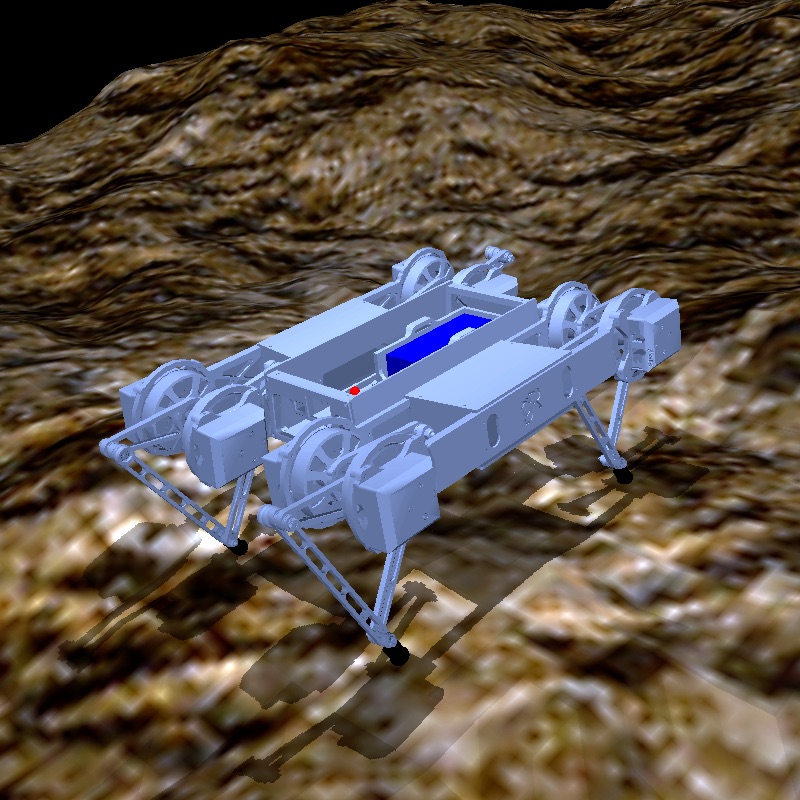}
\caption{Minitaur}
\label{fig:minitaur_sim}
\end{subfigure}
\begin{subfigure}[t]{0.24\columnwidth}
\includegraphics[width=\textwidth]{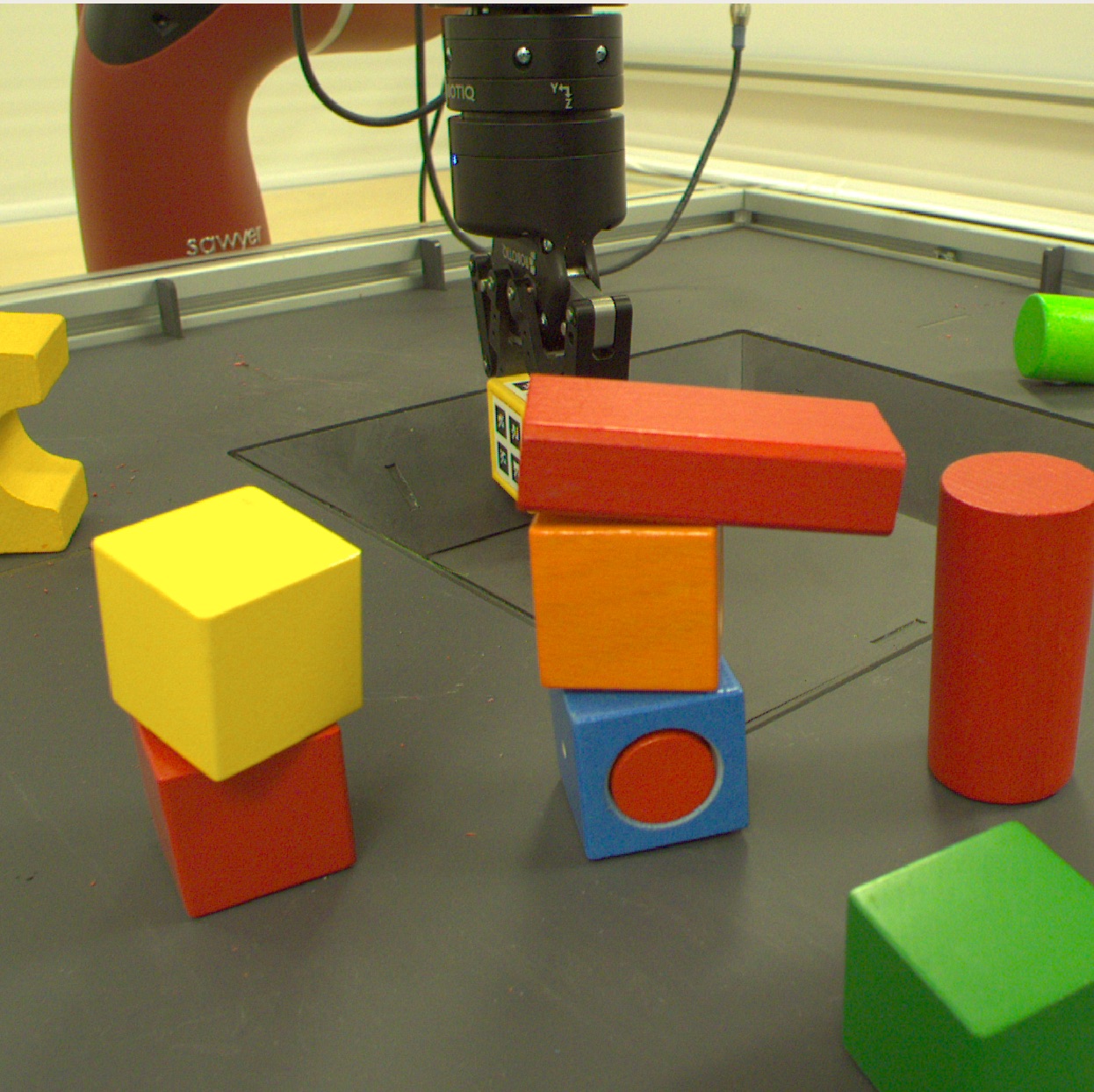}
\caption{Sawyer}
\label{fig:sawyer}
\end{subfigure}
\caption{\small{The continuous control environments used in the experiments.
Cart-pole swingup (\subref{fig:cartpole_sim}) and humanoid stand and walk (\subref{fig:humanoid_sim}) are from the DM control suite \citep{Tassa2018ControlSuite}.
The Minitaur robot (\subref{fig:minitaur_sim}) is similarly simulated in MuJoCo.
Finally, we use a real Sawyer robot (\subref{fig:sawyer}) for a reaching task.}}
\label{fig:environments}
\end{figure}

\section{Experiments}
\label{sec:experiments}

In order to understand the generality and potential impact of our approach, we experiment in the four continuous control domains shown in \Figref{fig:environments}: the cart-pole and humanoid from the DM Control Suite benchmark, a more challenging, realistically-simulated robot locomotion task, and finally two variants of a reaching task on a real robot arm.

\subsection{Control benchmarks}
\label{sec:cartpole}

We consider three tasks from the DeepMind Control Suite~\citep{Tassa2018ControlSuite} benchmark to illustrate the problem of bang-bang control specifically, and test the effectiveness of our approach: \emph{cart-pole swingup}, \emph{humanoid stand} and \emph{humanoid walk}.
Each of these tasks, by default, has a shaped reward that combines the success criterion (e.g. pole upright and cart in the center for cart-pole) with a bonus for a low control signal.
The total reward lies in $\left[0,1\right]$ in all cases.
We compare agents trained on this original reward with two variants: i) an agent trained with the control term from the reward removed, ii) an agent trained without the control term but using the proposed approach from \Secref{sec:approach}.
In all cases we learn a neural network controller using the \gls{mpo} algorithm~\citep{abdolmaleki2018maximum}.
More specifically, we train a two-layer MLP policy to output the mean and variance of a Gaussian policy.
For the constrained optimization approach, we use a fixed lower bound on the expected per-step reward of $0.9$ and use the norm of the force output as the penalty.
More details about the training setup can be found in Appendix A.
\Tabref{tab:suite} shows the average reward (excl. control penalty) and control penalty for each of the tasks and setups, both averaged across the entire episode as well as the final 50\%.
The latter is relevant as all three tasks have an initial balancing component, that by its nature requires significant control input.

\begin{table}[t!]
\small
\caption{\small{Average (reward / penalty) for the different control benchmark tasks and policies trained in the \bf{constrained}, \bf{unconstrained} and \bf{original} reward setup, and with reward and penalty computed over the \bf{full} episode, or only the last \bf{half}.
In all cases, the constraint-based approach results in the lowest average penalty (green is low and red is high).
While the lower bound was set to $0.9$ of a maximum of $1$, we obtain the same average reward as the unconstrained case for the cart-pole swingup and humanoid stand tasks.
}}
\label{tab:suite}
\centering
\begin{tabular}{cc|l|l|l}
\multicolumn{1}{c}{\bf Task} &\multicolumn{1}{c}{\bf Win.} &\multicolumn{1}{c}{\bf Constrained} &\multicolumn{1}{c}{\bf Unconstr.} &\multicolumn{1}{c}{\bf Original}\\
\hline
\multirow{2}{*}{cartpole} & full & 0.891 / \green{0.302} & 0.885 / \red{1.918} & 0.895 / 0.733\\
& last & 0.998 / \green{0.013} & 1.000 / \red{1.459} & 0.998 / 0.074\\[1mm]
human. & full & 0.961 / \green{5.608} & 0.964 / \red{37.19} & 0.952 / 27.52\\
(stand) & last & 0.998 / \green{4.538} & 0.993 / \red{37.29} & 0.999 / 27.01\\[1mm]
human. & full & 0.869 / \green{21.60} & 0.953 / 26.84 & 0.957 / \red{29.57}\\
(walk) & last & 0.903 / \green{21.60} & 0.984 / 26.82 & 0.990 / \red{29.42}\\
\end{tabular}
\end{table}

\begin{figure*}[t!]
\centering
\begin{subfigure}{0.48\textwidth}
\includegraphics[width=\textwidth]{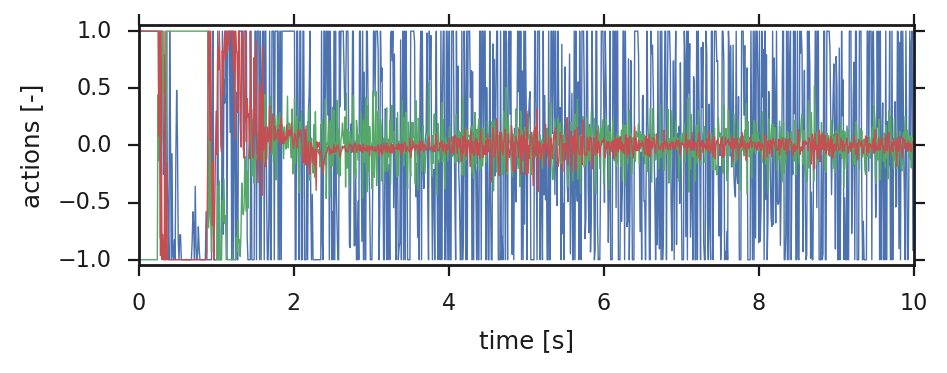}
\caption{Cart-pole}
\label{fig:cartpole_comparison}
\end{subfigure}
\begin{subfigure}{0.48\textwidth}
\includegraphics[width=\textwidth]{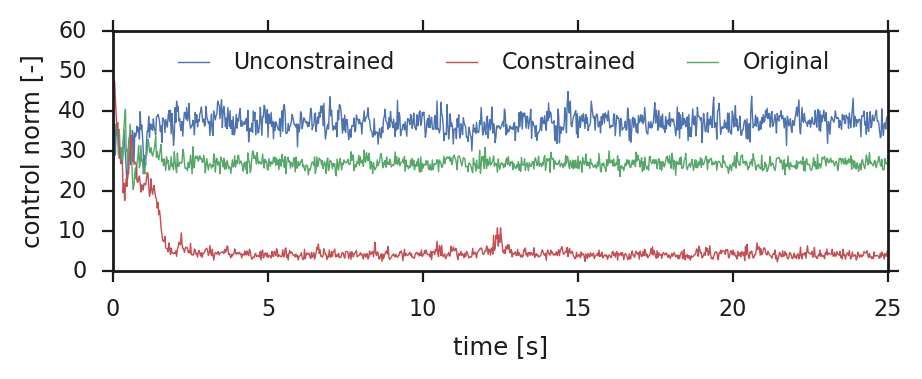}
\caption{Humanoid (stand)}
\label{fig:humanoid_stand_comparison}
\end{subfigure}
\caption{\small{Representative results of the executed policies in the control benchmark tasks.
Plot (\subref{fig:cartpole_comparison}) shows a representative rollout of the (1-dimensional) policy trained on cart-pole swingup in the unconstrained, constrained and original reward setting.
In all three cases, we observe high control input during the first 2 seconds, corresponding to the swingup phase.
Figure (\subref{fig:humanoid_stand_comparison}) shows the control norm during the episode rollout of policies trained in humanoid stand.
Note that in both tasks the actual return between the thee methods is almost identical.
}}
\label{fig:cartpole_results}
\end{figure*}

For cart-pole, we see that all agents obtain almost identical returns.
The constrained method, however, is able to achieve significantly lower penalties, even compared to the original reward that included a (non-adaptive) penalty.
\Figref{fig:cartpole_comparison} shows a comparison of the typical rollout of the different optimization strategies.
When optimizing for the reward alone, we can observe that the average absolute control signal is large and the agent keeps switching rapidly between a large negative and large positive force.
While the agent is able to solve the task (and the behaviour can be somewhat smoothed by using the policy mean instead of sampling), this kind of \emph{bang-bang} control is not desirable for any real-world control system.
The policy learned with the constrained approach is visibly smoother; in particular it never reaches maximum or minimum actuation levels after the swing up (during which a switch between maximum and minimum actuation is indeed the optimal solution).
For the agent trained with the original reward function, which incorporates a fixed control penalty, the action distribution also shrinks after the swingup phase, but not as much as in the constraint-based approach.

\begin{figure}[tbh]
\centering
\begin{subfigure}{0.3\columnwidth}
\includegraphics[width=\textwidth]{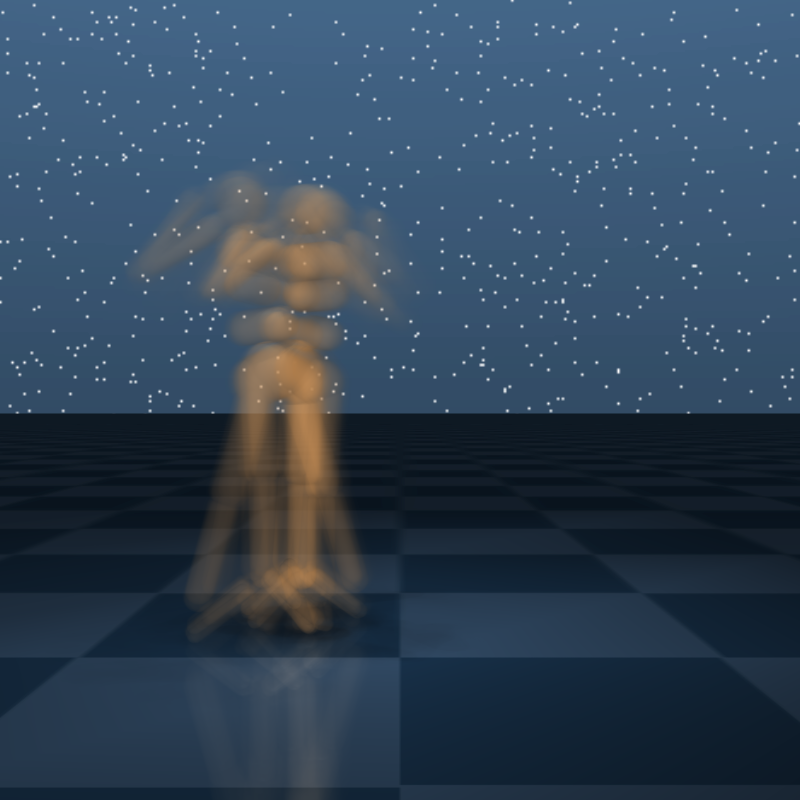}
\caption{Unconstrained}
\label{fig:humanoid_unconstrained}
\end{subfigure}
\hspace{0.02\columnwidth}
\begin{subfigure}{0.3\columnwidth}
\includegraphics[width=\textwidth]{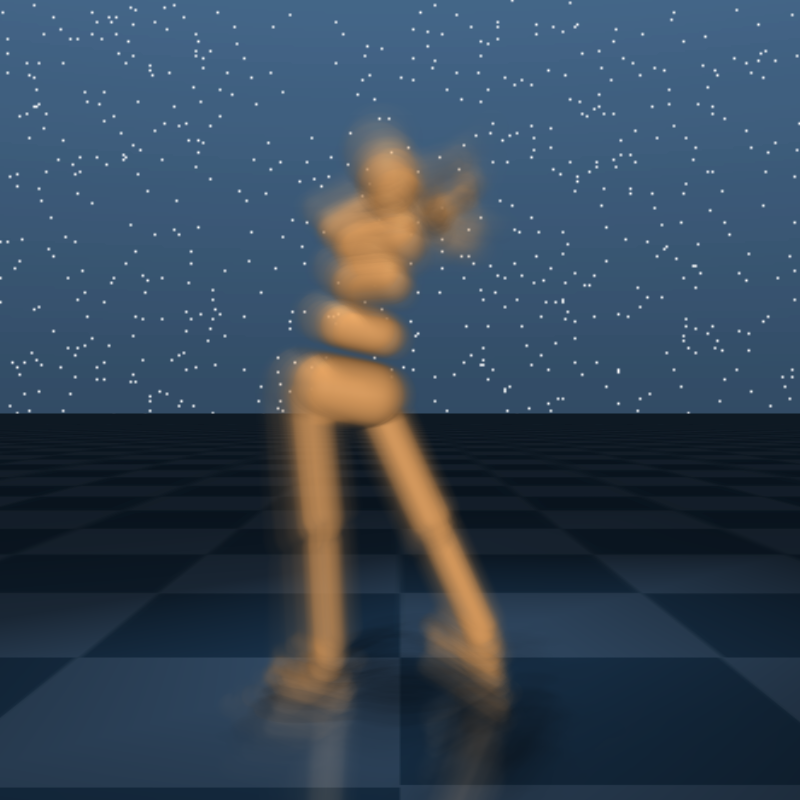}
\caption{Constrained}
\label{fig:humanoid_constrained}
\end{subfigure}
\hspace{0.02\columnwidth}
\begin{subfigure}{0.3\columnwidth}
\includegraphics[width=\textwidth]{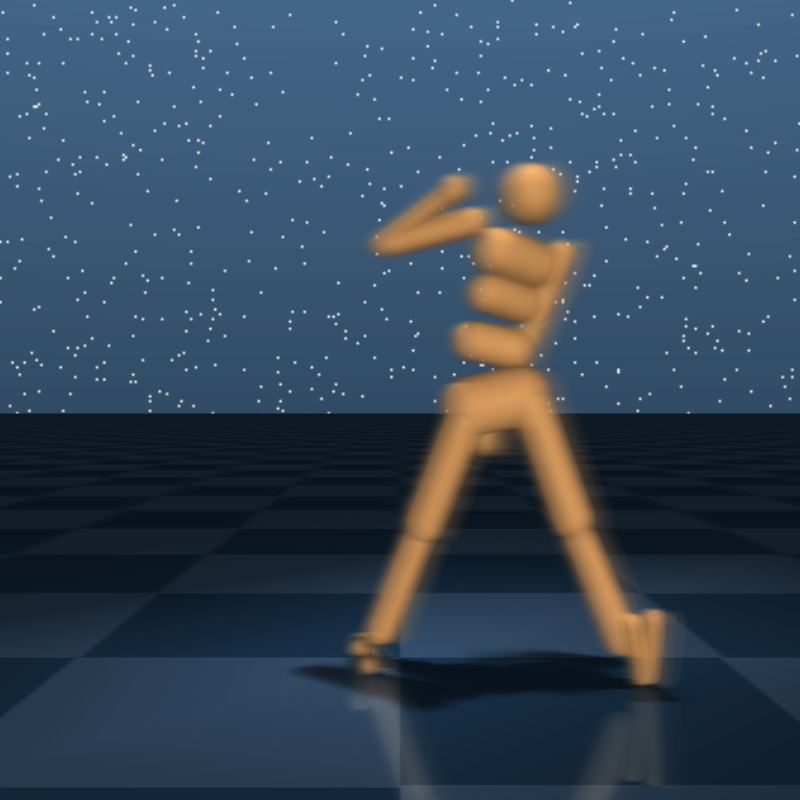}
\caption{Original}
\label{fig:humanoid_original}
\end{subfigure}
\caption{\small{Comparison of policies trained on the humanoid stand task in the constrained, unconstrained and original reward setup.
Figures show the average frame of the final 50\% of the episode.
Policies that exhibit more bang-bang-style control will result in more jittering movements and hence more blurry images.
}}
\label{fig:humanoid_results}
\end{figure}

We observe a similar trend for the humanoid stand task, where all three setups result in almost the same average reward, but the constraint-based approach is able to reduce the control penalty by 80\% compared to the original reward setup.
We visualize the resulting policies in \Figref{fig:humanoid_results} by overlaying frames from the final 50\% of the episode.
Bang-bang control will result in a more jittery motion and hence a more blurry image, as can be seen in \Figref{fig:humanoid_unconstrained}.
In contrast, both the constrained and original setup show a fixed pose and significantly less jitter.
In the constrained case, however, the agent consistently learn to use a smaller control norm by putting the legs closer together.
This can be observed in Figure~\ref{fig:humanoid_stand_comparison}, where, after the initial standup, the constrained optimization approach results in a lower control norm.

For the humanoid walk task, we observe that while the constraint-based approach still results in a lower penalty, there is also a reduction in the average reward.
This is to be expected: when walking, the penalty can be minimized by slowing down, thus the average per-step reward will stick closer to the imposed lower bound of $0.9$.
Interestingly, the original reward configuration results in a higher control penalty compared to the unconstrained case, perhaps because the control penalty is mixed into the reward differently than in the (un)constrained case and may hence result in a different optimum of the reward.

\subsection{Minitaur locomotion}
\label{sec:minitaur}

\begin{figure*}[t!]
\centering
\begin{subfigure}{0.22\textwidth}
\includegraphics[width=\textwidth]{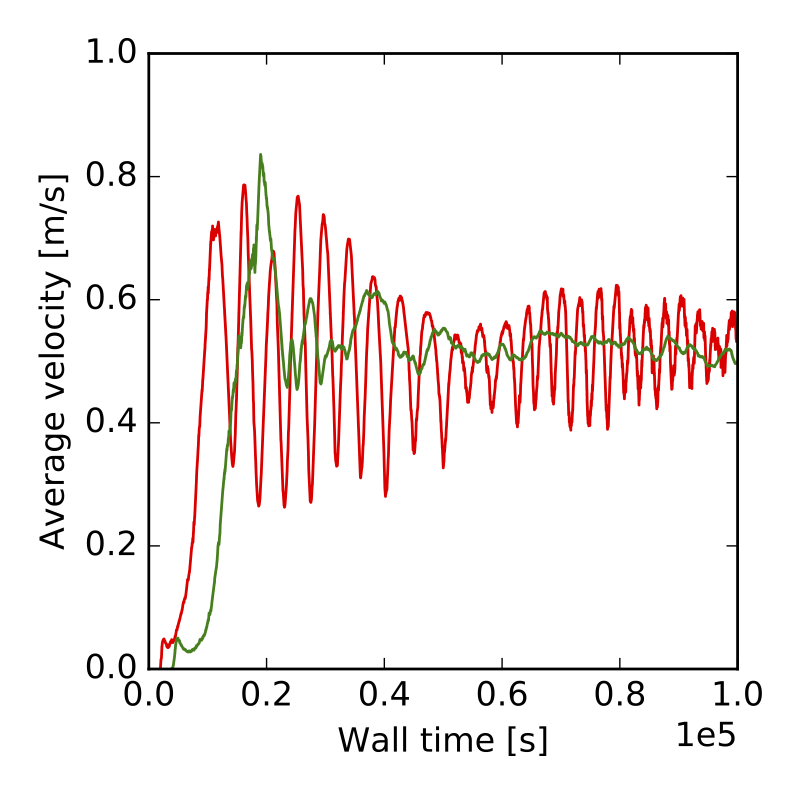}
\caption{}
\label{fig:minitaur_reward_time}
\end{subfigure}
\hspace{0.04\textwidth}
\begin{subfigure}{0.22\textwidth}
\includegraphics[width=\textwidth]{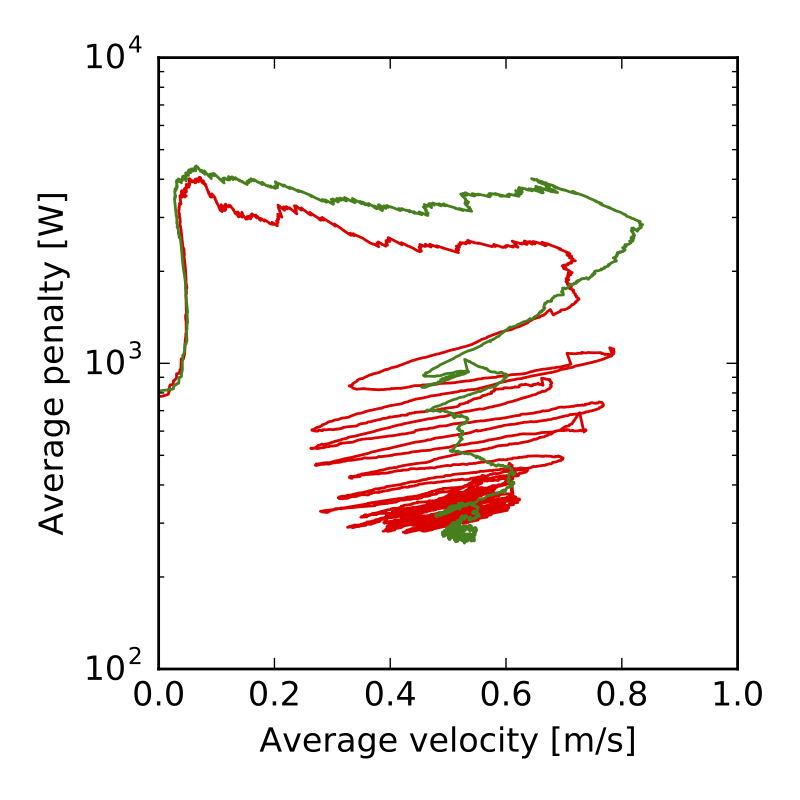}
\caption{}
\label{fig:minitaur_reward_penalty}
\end{subfigure}
\hspace{0.04\textwidth}
\begin{subfigure}{0.22\textwidth}
\includegraphics[width=\textwidth]{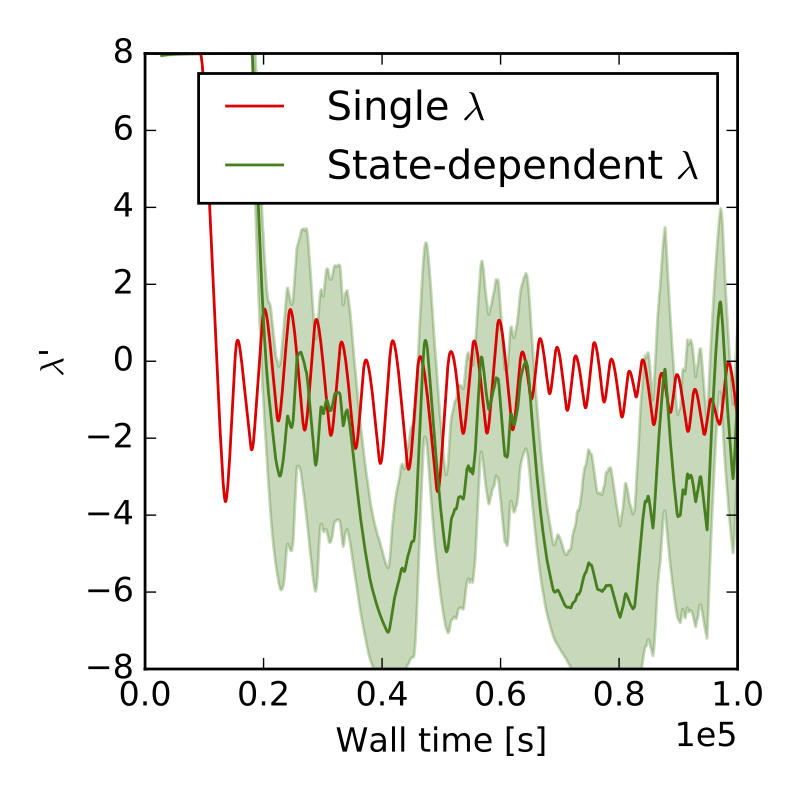}
\caption{}
\label{fig:minitaur_lambda_time}
\end{subfigure}
\caption{\small{
Comparison of a single versus a state-dependent $\lambda$ multiplier for models trained to achieve a minimum velocity of 0.5 m/s.
A single multiplier results in large swings in reward and on average higher values of $\lambda$.
(\subref{fig:minitaur_reward_time}) shows the per-step reward over time.
(\subref{fig:minitaur_reward_penalty}) shows the trade-off between the per-step reward and penalty during training.
Policies start off at 0 m/s and first learn to satisfy the constraint before optimizing the penalty.
(\subref{fig:minitaur_lambda_time}) shows the Lagrangian multiplier(s) change over time.
For the state-dependent case, we show the mean and standard deviation of $\lambda$ across the training batch.
}}
\label{fig:minitaur_training}
\end{figure*}

Our second simulated experiment is based on the Minitaur robot developed by Ghost Robotics \citep{Kenneally2016minitaur}.
The Minitaur is a quadruped with two \glspl{dof} in each of the four legs which are actuated by high-power direct-drive motors, allowing it to express various dynamic gaits such as trotting, pronking and galloping.
Implementing these gaits with state-of-the-art control techniques requires significant effort, however, and performance becomes sensitive to modeling errors when using model-based approaches.
Learning-based approaches have shown promise as an alternative for devising locomotion controllers~\citep{TanRSS2018}.
They are less dependent on gait and other task-dependent heuristics and can lead to more versatile and dynamic behaviors.
We propose learning gaits that are sufficiently smooth and efficient by optimizing for power usage.
This will avoid high-frequency changes in the control signal that in turn could cause  instability or mechanical stress.

Although the reported experiments are conducted in simulation, we have made a significant effort to capture many of the challenges of real robots.
We model the Minitaur in MuJoCo~\citep{Todorov2012Mujoco}, as depicted in \Figref{fig:minitaur_sim}, using model parameters obtained from data sheets as well as system identification to improve the fidelity.
The Minitaur is placed on a varying, rough terrain that is procedurally generated for every rollout.
We use a non-linear actuator model based on a general DC motor model and the torque-current characteristic described in \citet{Avik2015}.
The observations include noisy motor positions, yaw, pitch, roll, angular velocities and accelerometer readings, but no direct perception of the terrain, making the problem partially observed.
The policy outputs position setpoints at 100Hz that are fed to a proportional position controller, with a delay of 20ms between sensor readings and the corresponding control signal, to match delays observed on the real hardware.
To improve robustness, and with the aim of simulation-to-real transfer, we perform domain randomization \citep{Tobin2017Randomization} on a number of model parameters, as well as apply random external forces to the body (see Appendix B for details).

As we are only considering forward locomotion, we set the reward $r\left(\vs,\va\right)$ to be the forward velocity of the robot's base expressed in the world frame.
The cost $c\left(\vs,\va\right)$ is the total power usage of the motors according to the actuator model.
As the legs can collide with the main body, when giving the agent access to the full control range, a constant penalty is added to the power penalty during any self-collision.
We use a largely similar training setup as in \Secref{sec:cartpole}; however, since the episodes are 30 sec in length and observations are partial and noisy, the agent requires memory for effective state estimation, we thus add an LSTM~\citep{hochreiter1997long} layer to the model.
In addition to learning separate values for $Q_{r}\left(\vs,\va\right)$ and $Q_{c}\left(\vs,\va\right)$, we split up $Q_{c}\left(\vs,\va\right)$ into separate value functions for the power usage and collision penalty.
We also increase the number of actors to 100 to sample a larger number of domain variations more quickly.
More details can be found in Appendix A.

We first evaluate the effect of applying the lower bound to each individual state instead of to the global average velocity.
\Figref{fig:minitaur_training} shows a comparison between the learning dynamics of a model using a single $\lambda$ multiplier and a model with a state-dependent one, i.e. constrained in expectation or per-step.
Both agents try to achieve a lower bound on the value that is equivalent to a minimum velocity of 0.5 m/s.
At first, both agents ``focus'' on satisfying the constraint, increasing the penalty significantly in order to do so.
Once the target velocity is exceeded, the agents start to optimize the penalty, which drives them back to the imposed bound.
We see that a single global $\lambda$ multiplier leads to large oscillations between moving too slow at a lower penalty and moving too fast at a higher penalty.
Although this process eventually converges, it is inefficient.
In contrast to this, the agent with the state-dependent $\lambda$ tracks the target velocity more closely, and achieves slightly lower penalties.
The state-dependent $\lambda$ shows generally lower values over time as well (\Figref{fig:minitaur_lambda_time}).

\begin{table*}[t!]
\small
\caption{\small{Results for models trained to achieve a fixed lower bound on the velocity.
Reported numbers are average per-step delta (velocity overshoot [m/s]) and penalty ([W]), except for the unbounded case where we report actual velocity.
Each entry is an average over 4 seeds.
We highlight the best constant $\alpha$, in terms of smallest overshoot, for each target bound.
The constraint version achieves overshoot comparable or smaller than the fixed alpha in each condition while achieving significantly lower penalty (coloring: green (good); red (bad)).
}}
\label{tab:minitaur_fixed}
\centering
\begin{tabular}{c|ll|ll|ll|ll|ll}
\multicolumn{1}{c}{\bf Target} &\multicolumn{2}{c}{\bf $\alpha=\expnumber{3}{-3}$} &\multicolumn{2}{c}{\bf $\alpha=\expnumber{1}{-3}$} &\multicolumn{2}{c}{\bf $\alpha=\expnumber{3}{-4}$} &\multicolumn{2}{c}{\bf $\alpha=\expnumber{1}{-4}$} &\multicolumn{2}{c}{\bf Constrained}\\
& delta & penalty & delta & penalty & delta & penalty & delta & penalty & delta & penalty \\
\hline
0.1 &-0.1, &35.74 &\textbf{-0.01}, &\textcolor{green}{\textbf{104.2}} &0.07, &112.35 &0.1, &245.49 &\textbf{0.01}, &\textcolor{red}{\textbf{127.14}}\\
0.2 &-0.2, &46.48 &\textbf{-0.01}, &\textcolor{red}{\textbf{210.04}} &0.15, &207.19 &0.23, &399.83 &\textbf{0.03}, &\textcolor{green}{\textbf{106.88}}\\
0.3 &-0.3, &50.3 &\textbf{0.06}, &\textcolor{red}{\textbf{154.91}} &0.16, &213.1 &0.24, &429.6 &\textbf{0.04}, &\textcolor{green}{\textbf{89.97}}\\
0.4 &-0.4, &54.05 &\textbf{0.06}, &\textcolor{red}{\textbf{195.98}} &0.11, &306.1 &0.32, &627.66 &\textbf{0.05}, &\textcolor{green}{\textbf{132.97}}\\
0.5 &-0.5, &60.71 &\textbf{0.13}, &\textcolor{red}{\textbf{250.69}} &\textbf{0.13}, &\textcolor{red}{\textbf{332.53}} &0.26, &808.38 &\textbf{0.05}, &\textcolor{green}{\textbf{142.93}}\\
$\infty$ &\textit{0.0}, &54.63 &\textit{1.25}, &775.08 &\textit{1.24}, &1556.97 &\textit{1.24}, &1656.42 &-, &-\\
\end{tabular}
\end{table*}

\begin{table*}[t!]
\small
\caption{\small{Results of models that are conditioned on the target velocity, evaluated for for different values.
Reported numbers are average per-step (velocity overshoot [m/s], penalty [W]).
Each row is an average over 4 seeds.
The highlighted numbers mark the best individual alpha for each target velocity (in terms of velocity overshoot).
As can be observed no single $\alpha$ performs well across target velocities.
In contrast the constraint version achieves low overshoot in all conditions; and also achieves lower penalty than the best $\alpha$ in all but one case (as indicated by the coloring: green (good) and red (bad)).
}}
\label{tab:minitaur_varying}
\centering
\begin{tabular}{c|ll|ll|ll|ll|ll}
\multicolumn{1}{c}{\bf Target} &\multicolumn{2}{c}{\bf $\alpha=\expnumber{3}{-3}$} &\multicolumn{2}{c}{\bf $\alpha=\expnumber{1}{-3}$} &\multicolumn{2}{c}{\bf $\alpha=\expnumber{3}{-4}$} &\multicolumn{2}{c}{\bf $\alpha=\expnumber{1}{-4}$} &\multicolumn{2}{c}{\bf Constrained}\\
& delta & penalty & delta & penalty & delta & penalty & delta & penalty & delta & penalty \\
\hline
0.0 & \textbf{0.0}, &\textcolor{green}{\textbf{53.68}} &0.01, &116.59 &0.17, &272.45 &0.37, &757.53 &\textbf{0.0}, &\textcolor{red}{\textbf{84.07}}\\
0.1 &-0.1, &54.49 &\textbf{0.0}, &\textcolor{red}{\textbf{158.68}} &0.21, &324.16 &0.37, &619.3 &\textbf{0.0}, &\textcolor{green}{\textbf{141.86}}\\
0.2 &-0.2, &53.54 &\textbf{0.02}, &\textcolor{red}{\textbf{256.68}} &0.21, &373.13 &0.36, &627.19 &\textbf{0.04}, &\textcolor{green}{\textbf{174.79}}\\
0.3 &-0.3, &53.6 &\textbf{-0.02}, &\textcolor{red}{\textbf{314.71}} &0.16, &336.48 &0.42, &747.24 &\textbf{0.02}, &\textcolor{green}{\textbf{188.18}}\\
0.4 &-0.4, &54.82 &\textbf{-0.07}, &\textcolor{red}{\textbf{384.94}} &0.15, &467.21 &0.32, &870.34 &\textbf{0.05}, &\textcolor{green}{\textbf{252.54}}\\
0.5 &-0.5, &52.37 &-0.1, &366.48 &\textbf{0.01}, &\textcolor{red}{\textbf{594.36}} &0.27, &1026.3 &\textbf{0.05}, &\textcolor{green}{\textbf{361.16}}\\
0.6 &-0.6, &52.36 &-0.2, &686.36 &-0.07, &770.67 &\textbf{0.02}, &\textcolor{red}{\textbf{1632.96}} &\textbf{-0.04}, &\textcolor{green}{\textbf{773.79}}\\
\end{tabular}
\end{table*}

In \Tabref{tab:minitaur_fixed}, we compare the reward-penalty trade-off of our approach to baselines where we clip the reward,  $r'\left(\vs_{t},\va_{t}\right)=\min\left(r\left(\vs_{t},\va_{t}\right),\target{r}\right)$, and use a fixed coefficient $\alpha$ for the penalty.
As there is less incentive for the agent to increase the reward over $\target{r}$, there is more opportunity to optimize the penalty.
Results shown are the per-step overshoot with respect to the desired target velocity and the penalty, averaged across 4 seeds and 100 episodes each (the first 100 ms are clipped to disregard transient behavior when starting from a stand-still).
We also compare to a baseline where the reward is unbounded, marked as $\infty$ in \Tabref{tab:minitaur_fixed}.
In the unbounded reward case, it proves to be difficult to achieve a positive but moderately slow speed.
Either $\alpha$ is too high and the agent is biased towards standing still, or it is too low and the agent reaches the end of the course before the time limit (corresponding to an average velocity of approx. 1.25 m/s).
For the clipped reward, we observe a similar issue when $\alpha$ is set too high.
In nearly all other cases, the targeted speed is exceeded by some margin that increases with decreasing $\alpha$.
While there is less incentive to exceed $\target{r}$, a larger margin decreases the chances of the actual speed momentarily dropping below the target speed.
Using the constraint-based approach, we generally achieve average actual speeds closer to the target and at a lower average penalty, showing the merits of adaptively trading of reward and cost.

\Tabref{tab:minitaur_varying} shows a comparison between agents trained across varying target speeds (sampled uniformly in $\left[0,0.5\right]$ m/s).
These agents are given the target speed as observations.
The evaluation procedure is the same as before, except that we evaluate the same conditional policy for multiple target values.
We make similar observations: a fixed penalty coefficient generally leads to higher speeds then the set target, and higher penalties.
Interestingly, for higher target velocities, the actual velocity exceeds the target less, indicating that different values for $\alpha$ are required for different targets.
As we learn multipliers that are conditioned on the target, we can track the target more closely, even for higher speeds.
We also evaluate these models for a target speed outside out the training range.
Performance degrades quite rapidly, with the constraint no longer satisfied, and at significantly higher cost.
This can be explained by the way the policies change behavior to match the target speed: generally the speed is changed by modulating the stride length.
Increasing the stride length much further than observed during training, however, results in collisions occurring that were not present at lower speeds, and hence higher penalties.
The same observation also explains why the penalties in the conditional case are higher than in the fixed case (final column in \Tabref{tab:minitaur_varying} vs. \Tabref{tab:minitaur_fixed}), as distinct behaviors are optimal for different target velocities.
This is likely a limitation of the relatively simple policy architecture, and improving diversity across goal velocities will be studied in future work.

\begin{figure*}[t!]
\centering
\begin{subfigure}{0.216\textwidth}
\includegraphics[width=\textwidth, trim=0 0 0 0.2cm]{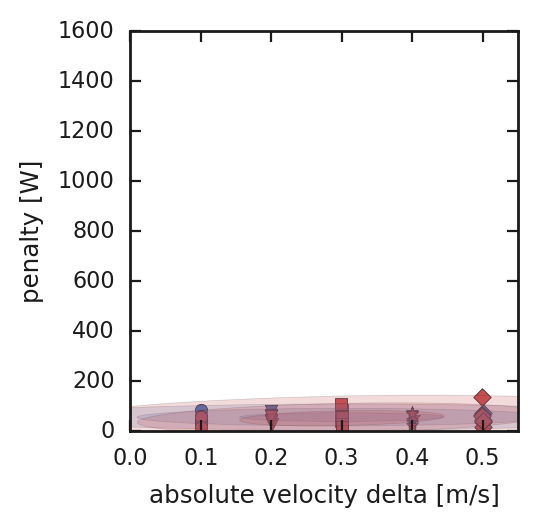}
\caption{$\alpha=\expnumber{3}{-3}$}
\label{fig:minitaur_alpha0003}
\end{subfigure}
\begin{subfigure}{0.18\textwidth}
\includegraphics[width=\textwidth]{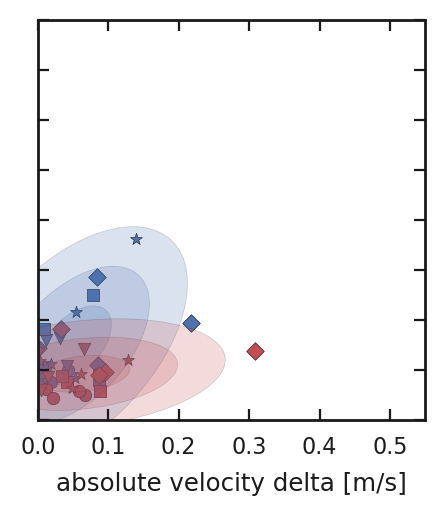}
\caption{$\alpha=\expnumber{1}{-3}$}
\label{fig:minitaur_alpha0001}
\end{subfigure}
\begin{subfigure}{0.18\textwidth}
\includegraphics[width=\textwidth]{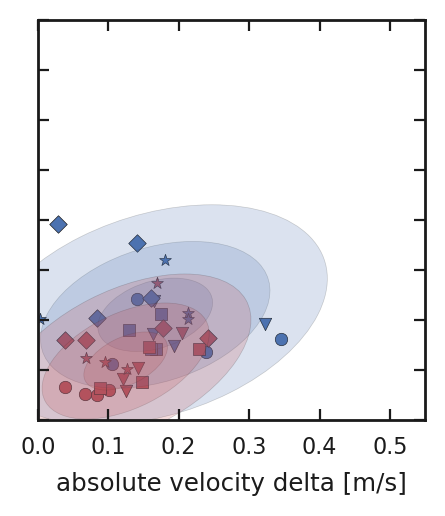}
\caption{$\alpha=\expnumber{3}{-4}$}
\label{fig:minitaur_alpha00003}
\end{subfigure}
\begin{subfigure}{0.18\textwidth}
\includegraphics[width=\textwidth]{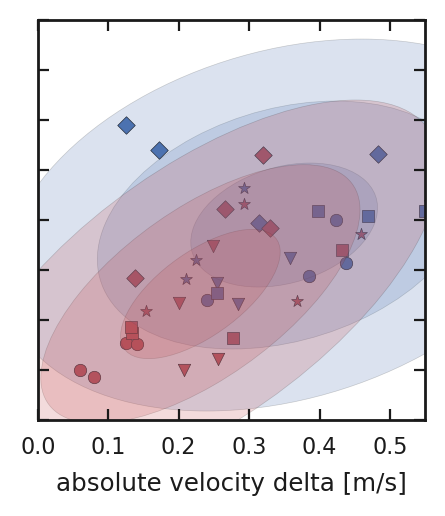}
\caption{$\alpha=\expnumber{1}{-4}$}
\label{fig:minitaur_alpha00001}
\end{subfigure}
\begin{subfigure}{0.18\textwidth}
\includegraphics[width=\textwidth]{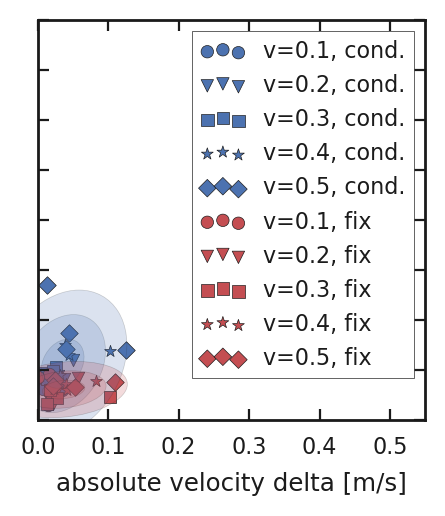}
\caption{Constrained}
\label{fig:minitaur_constraint}
\end{subfigure}
\caption{\small{Comparison of the constrained optimization approach with baselines using a fixed penalty.
Each data point shows the average absolute velocity delta and penalty for an agent optimized for a specific target velocity.
The different ellipse shades show one to three standard deviations, both for the fixed (red) and the varying (blue) velocity setpoints.
For each setting we train four agents.
In the fixed target case, these are different models.
In the conditional target case, these are evaluations of a single model conditioned on desired velocities.
}}
\label{fig:minitaurcomparison}
\end{figure*}

\Figref{fig:minitaurcomparison} extends the comparisons by plotting penalty over absolute velocity deltas for the different approaches.
The plots show that finding a suitable weighting that works for all tasks and setpoints is difficult.
While it is easy to identify values for $\alpha$ that are clearly too high or low, performance over tasks can vary even for well-tuned values.
Our approach as shown in \Figref{fig:minitaur_constraint} achieves a consistent performance, with low velocity overshoot errors and low penalty across all tests.
These results suggest that since our approach is less sensitive to task-specific changes, it may also greatly reduce computationally expensive hyperparameter tuning.
Videos showing some of the learned behaviors, both in the fixed and conditional constraint case, can be found at \url{https://sites.google.com/view/successatanycost}.

\subsection{Sawyer reaching with visibility constraint}
\label{sec:sawyer}

\begin{figure*}[t!]
\centering
\begin{subfigure}{0.16\textwidth}
\includegraphics[width=\textwidth]{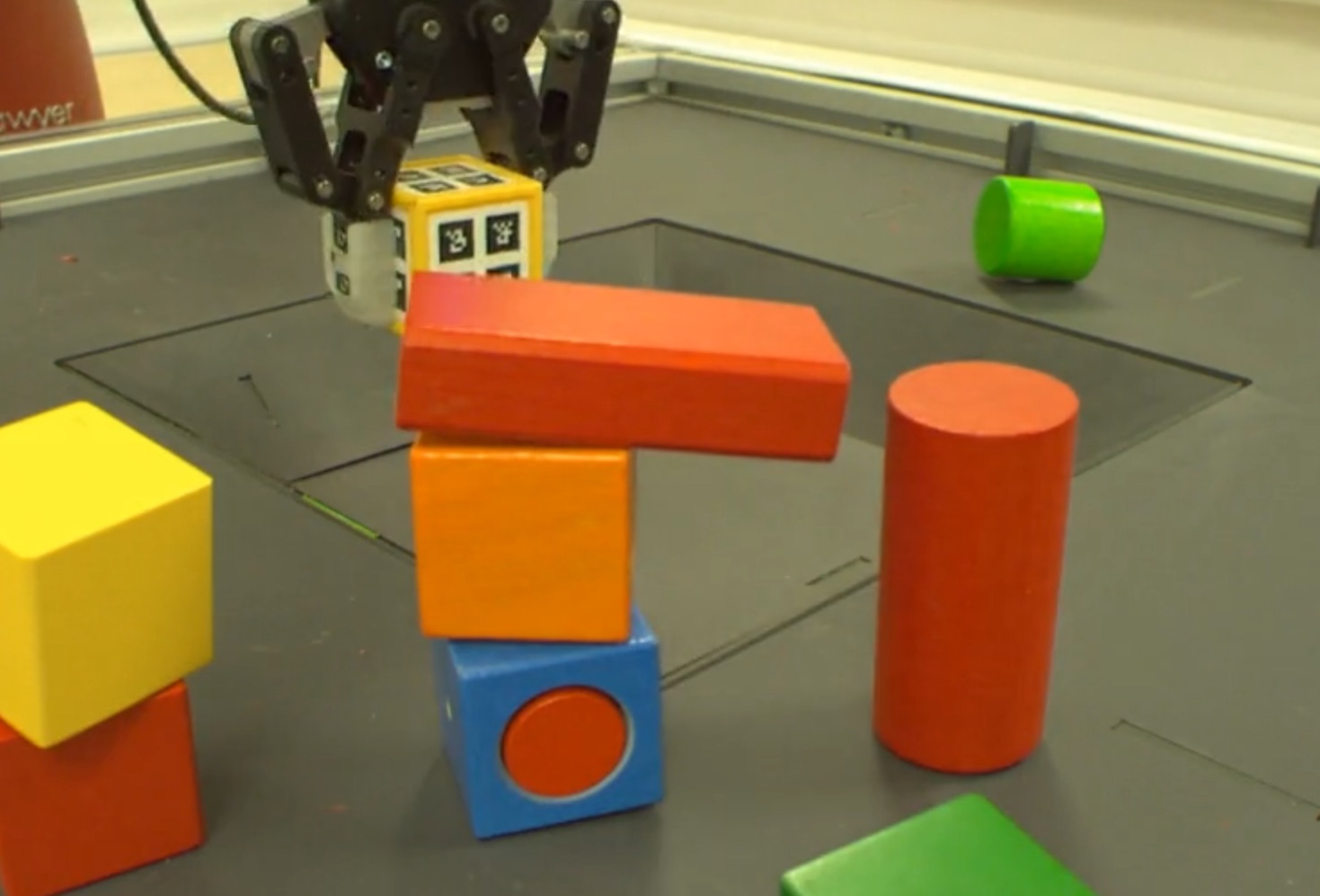}
\end{subfigure}
\begin{subfigure}{0.16\textwidth}
\includegraphics[width=\textwidth]{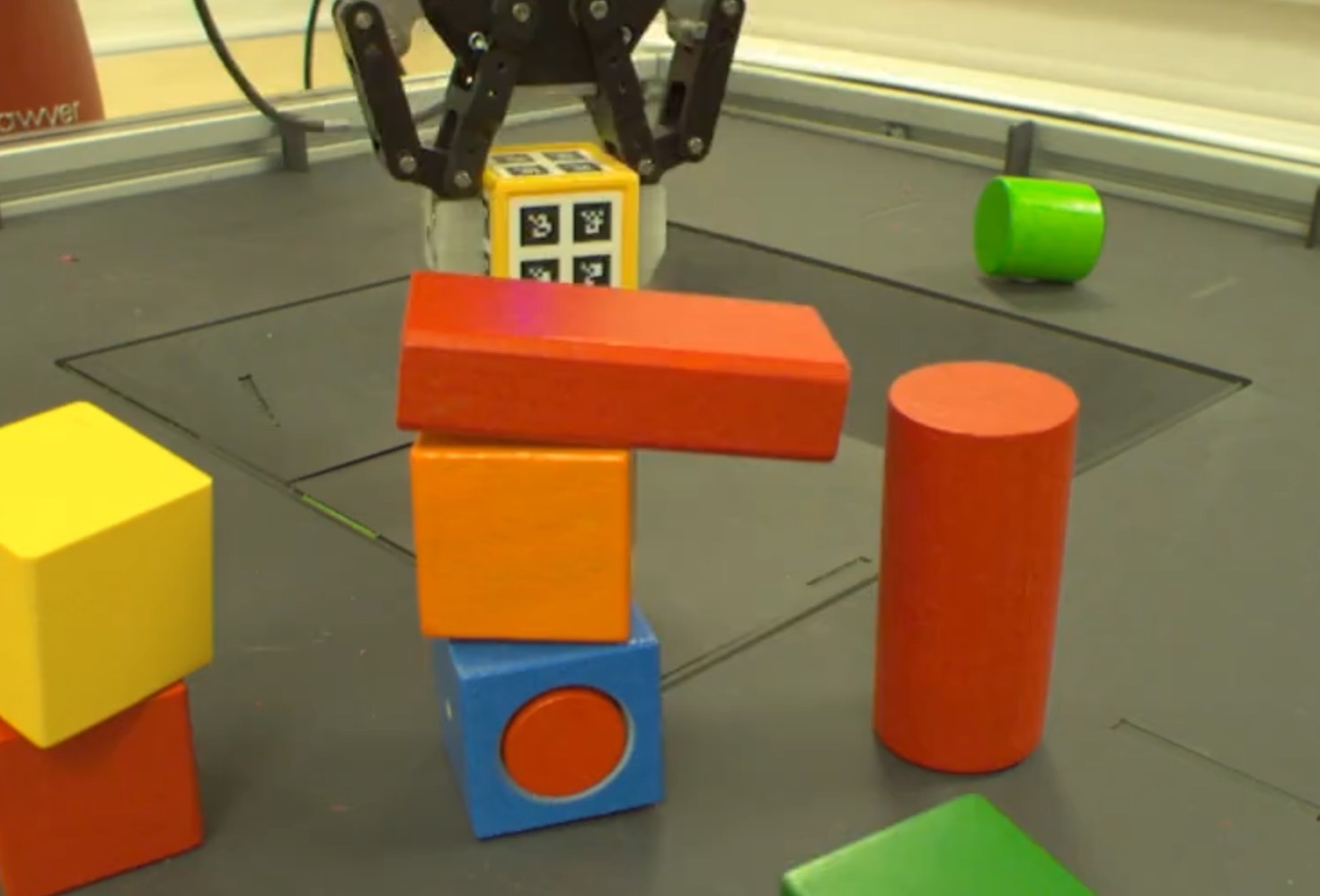}
\end{subfigure}
\begin{subfigure}{0.16\textwidth}
\includegraphics[width=\textwidth]{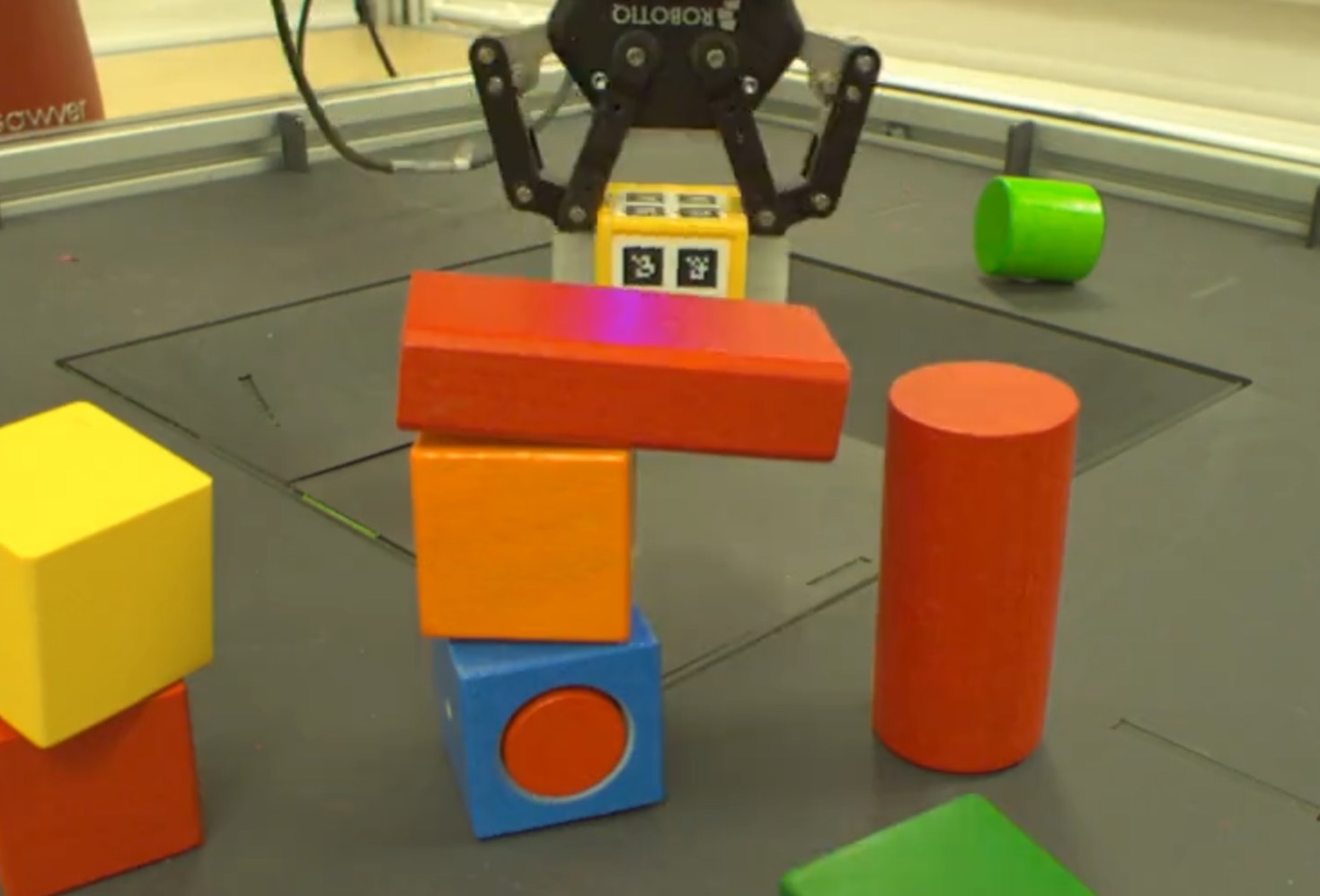}
\end{subfigure}
\begin{subfigure}{0.16\textwidth}
\includegraphics[width=\textwidth]{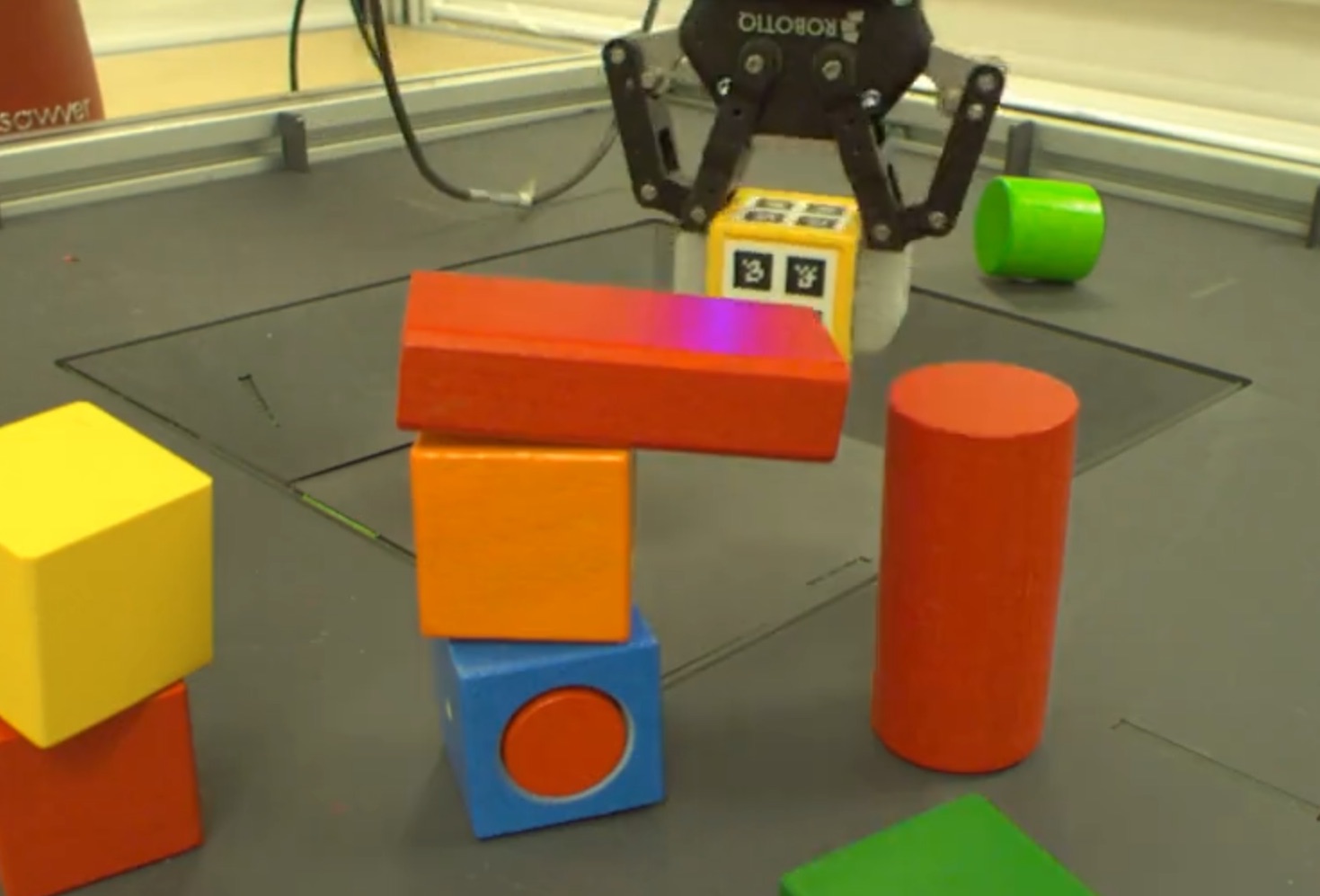}
\end{subfigure}
\begin{subfigure}{0.16\textwidth}
\includegraphics[width=\textwidth]{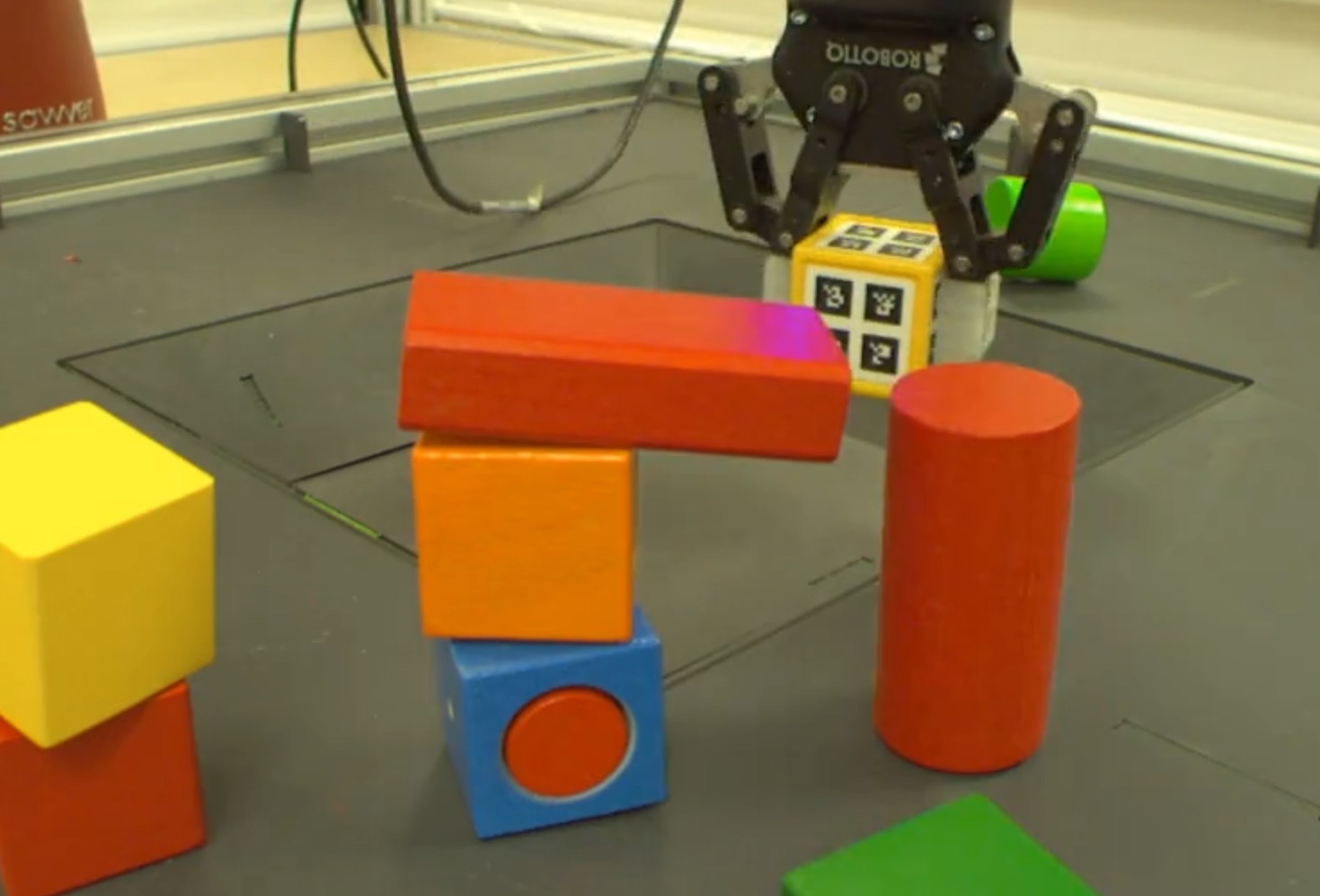}
\end{subfigure}
\begin{subfigure}{0.16\textwidth}
\includegraphics[width=\textwidth]{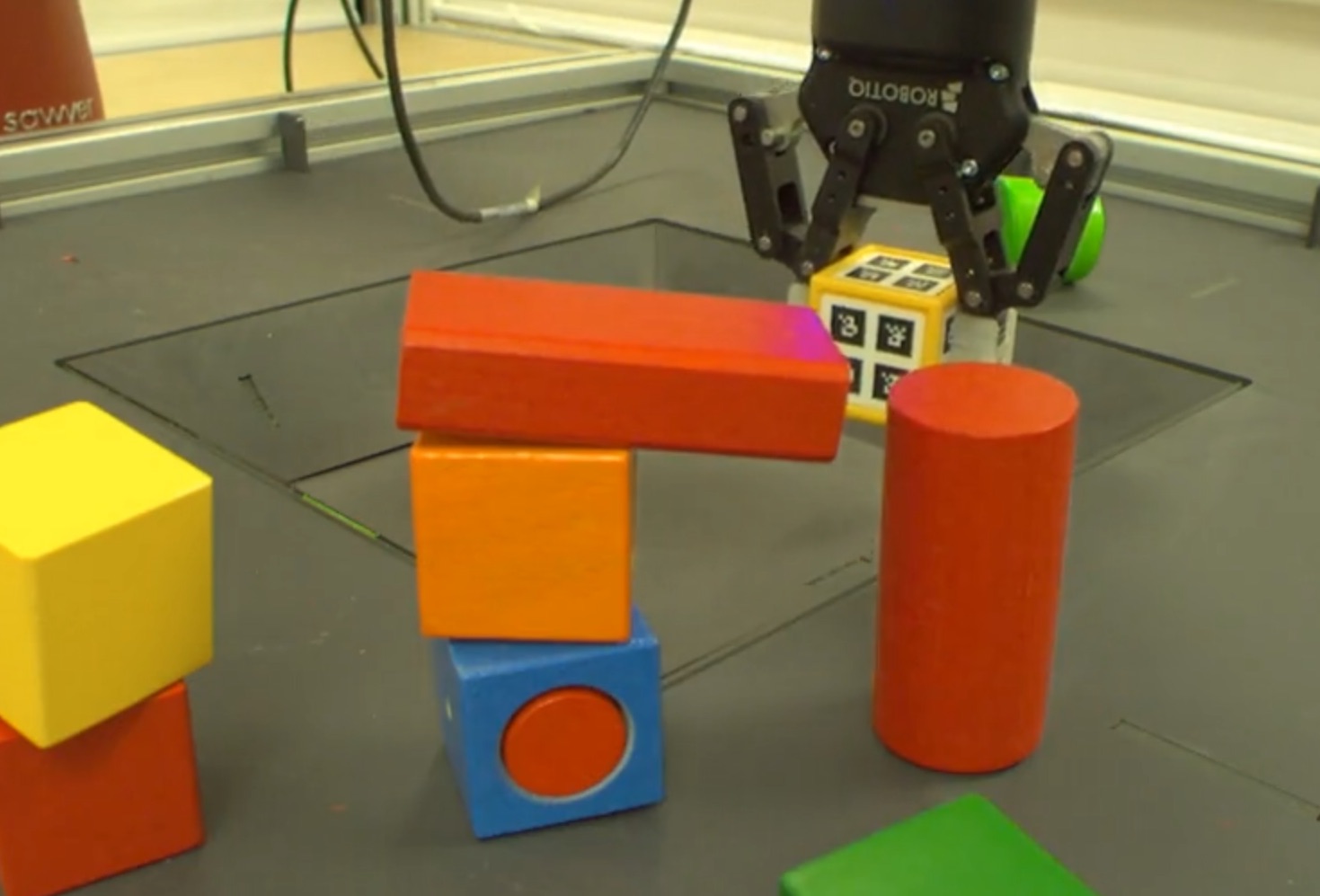}
\end{subfigure}
\caption{\small{Time series of a test rollout of the reaching task with the visibility constrained.
The time series shows three distinct behaviours the agent has learned: it follows the contour of the obstacle during the reach, ensuring visibility of the top two tags.
Additionally, it rotates the wrist to keep the cube facing the camera.
And lastly, when reached close to the target, it keeps a single tag visible in between two objects.
}}
\label{fig:sawyer_timeseries}
\end{figure*}

To demonstrate that our algorithm can, without modification, be used on robotic hardware we apply it to a reaching task on a robot arm in a crowded tabletop environment.
To explore the versatility of the constraint-based approach, we design the task such that it contains a reward objective, as well as a constraint.
The agent must learn to reach to a random 3D target location while maintaining constant visibility. 

In more detail, the robot is a Sawyer 7 DoF arm mounted on a table and equipped with a Robotiq 2F-85 parallel gripper.
We place a 5~cm wide cube inside the gripper, and track the cube with a camera using fiducials (augmented reality tags).
The objective is to reach a virtual target position sampled within a 20 cm cube workspace.
A number of obstacles are however placed in front of the camera, as seen in \Figref{fig:sawyer_timeseries}.
There are two ways the agent can lose visibility: either the cube is occluded by obstacles or the wrist is rotated such that the cube faces away from the camera.
Hence, we add an objective to keep the cube visible and constrain it by a lower bound; the visibility is a binary signal indicating whether at least one of the cube's fiducials is detected in the camera frame.

To phrase this in the framework of \Secref{sec:approach}, the reward in this case is the visibility, which we constrain to be true 95\% of the time.
The negative cost is now a sigmoidal function of the distance of the cube to the target, being at most 1 and decaying to 0.05 over a distance of 10 cm.

The policy and critic receive several inputs: the proprioception (joint positions, velocities and torques), the visibility indicator, and the previous action taken by the agent.
We use an action and observation history of 2, i.e. the past two observations/actions are provided to the agent.
The policy outputs 4-dimensional Cartesian velocities: three translational degrees of freedom (limited to [-0.07, 0.07] m/s) plus wrist rotation (limited to [-1, 1] rad/s).
The policy is executed at a 20 Hz control rate and we limit each episode to 600 steps.
As the camera image itself is not observed, the obstacle configuration has to be indirectly inferred through trial and error.
We hence keep the obstacle configuration fixed during training, though the setup can be extended to varying obstacle configurations if the camera image is also observed.

This task setup shows a possible application of the proposed method to a more complex task.
Other approaches to solving such a task exist (weighted costs, multiplicative costs or early episode termination).
However, the problem formulation of a hard constrained performance metric with a secondary reward objective that is compatible with the constraint (i.e. being defined in the nullspace of the constraint) feels very natural in our approach.

\begin{figure}[tbh]
\centering
\begin{subfigure}{0.49\columnwidth}
\includegraphics[width=\textwidth]{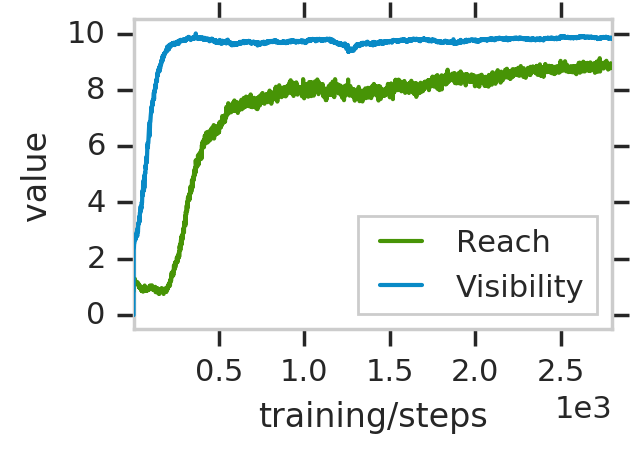}
\caption{\small{Values over time}}
\label{fig:sawyer_reach_values}
\end{subfigure}
\begin{subfigure}{0.49\columnwidth}
\includegraphics[width=\textwidth]{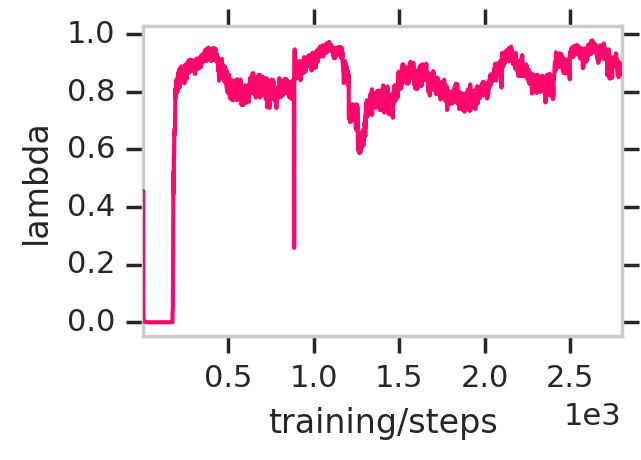}
\caption{\small{Multiplier over time}}
\label{fig:sawyer_reach_lambda}
\end{subfigure}
\caption{\small{
Learning progress of the Sawyer reaching task with visibility constraint.
(\subref{fig:sawyer_reach_values}) shows the evolution of the reaching and visibility values over time.
(\subref{fig:sawyer_reach_lambda}) shows the evolution of the ratio of the reaching objective vs. visibility objective, as averaged over the sampled batch.
Learning focuses solely on the visibility objective for the first 200 steps.
}}
\label{fig:sawyer_occlusion}
\end{figure}

\Figref{fig:sawyer_occlusion} shows the learning progress of the agent in this task.
Taking a look at the value estimates, we see that optimization initially focuses on increasing the value of the visibility objective, while the value for the reaching objective does not change much.
Once the lower bound of 95\% visibility, corresponding to a value of 9.5, has been met after about 200 training steps, the value of the reaching task starts to increase as well.
After about 2500 steps the reaching objective has also achieved its maximum achievable return (without violating the constraint).
The visibility value remains nearly constant during the remainder of learning.
We observe the same trend in \Figref{fig:sawyer_reach_lambda}, which shows the ratio, as defined by \Eqref{eq:normalisation}, of the reaching objective versus the visibility objective.
Initially all the weight is put on the visibility objective, which then shifts to about 80\% reaching and 20\% visibility.
Note that the average ratio is plotted, but the actual ratio will differ for each state encountered by the policy.
Figure~\ref{fig:sawyer_timeseries} shows an example rollout of the learned policy.
The agent is able to avoid shortcuts that affect visibility, maintains a wrist rotation facing the camera and settling in a position where one tag remains visible at all times.
A video showing the behavior of the learned policy qualitatively can be found at \url{https://sites.google.com/view/successatanycost}.

\section{Conclusion}

In order to regularize behavior in continuous control \gls{rl} tasks in a controllable way, we introduced a constraint-based RL approach that is able to automatically trade off rewards and penalties, and can be used in conjunction with any model-free, value-based \gls{rl} algorithm.
The constraints are applied in a point-wise fashion, for each state that the learned policy encounters. 
The resulting constrained optimization problem is solved using Lagrangian relaxation by iteratively adapting a set of Lagrangian multipliers, one per state, during training.
We show that we can learn these multipliers in the critic model alongside the value estimates of the policy, and closely track the imposed bounds.
The policy and critic can furthermore generalize across lower bounds by making the constraint value observable, resulting in a single conditional \gls{rl} agent that is able to dynamically trade off reward and costs in a controllable way.
We applied our approach to a number of continuous control benchmarks and show that without some cost function, we observe high-amplitude and high-frequency control.
Our method is able to reduce the control actions significantly, sometimes without sacrificing average reward.
In a simulated quadruped locomotion task, we are able to minimize electrical power usage with respect to a lower bound on the forward velocity.
We show that our method can achieve both lower velocity overshoot as well as lower power usage compared to a baseline that uses a fixed penalty coefficient.
Finally, we successfully learn a reaching tasks in a cluttered tabletop environment on a real robot arm with a visibility constraint, demonstrating that our method extends to real world system and non-trivial problems.

\bibliographystyle{plainnat}
\bibliography{references}

\clearpage
\onecolumn

\section*{Appendix A: Optimization details}

\subsection{General algorithm}
The general outline of the optimization procedure for \Eqref{eq:constraint_loss} is listed in Algorithm~\ref{alg:general}.
The approach is compatible with any actor-critic algorithm; in the next paragraphs we detail the methods used in this paper for policy evaluation and optimization.

\begin{algorithm}[h]
\caption{\small{Value constrained model-free control}}
\label{alg:general}
\begin{algorithmic}[1]
\State{\bf given} 
$Q_{r}\left(\vs,\va;\vpsi_{r}^{\left(0\right)},\vphi^{\left(0\right)}\right)$,
$Q_{c}\left(\vs,\va;\vpsi_{c}^{\left(0\right)},\vphi^{\left(0\right)}\right)$,
$\lambda\left(\vs;\vpsi_{\lambda}^{\left(0\right)},\vphi^{\left(0\right)}\right)$,
$\pi(\va | \vs ; \vtheta^{\left(0\right)})$, with $\vpsi_{i}^{\left(0\right)}$, $\vphi^{\left(0\right)}$ and $\vtheta^{\left(0\right)}$ initial weights, and replay buffer $\mathcal{D}$
\Repeat
\State{Execute $\va \sim \pi(\va | \vs ; \vtheta^{\left(0\right)})$ and observe $\vs'$, $r\left(\vs,\va\right)$, $c\left(\vs, \va\right)$}
\State{Add tuple $\left(\vs, \va, \vs', r\left(\vs,\va\right), c\left(\vs, \va\right)\right)$ to $\mathcal{D}$}
\State{Sample batch $\mathcal{B}$ of tuples from $\mathcal{D}$}
\State{\bf Critic update:}
\State{$L_r\left(\vpsi_{r}^{\left(k\right)}, \phi^{\left(k\right)}\right) = \E_{\mathcal{B}}\left[\text{valueLoss}\left(\vs,\va,\vs',r\left(\vs,\va\right),Q_{r}\left(\vs,\va;\vpsi_{r}^{\left(k\right)},\vphi^{\left(k\right)}\right)\right)\right]$}
\State{$L_c\left(\vpsi_{c}^{\left(k\right)}, \phi^{\left(k\right)}\right) = \E_{\mathcal{B}}\left[\text{valueLoss}\left(\vs,\va,\vs',c\left(\vs,\va\right),Q_{c}\left(\vs,\va;\vpsi_{c}^{\left(k\right)},\vphi^{\left(k\right)}\right)\right)\right]$}
\State{$L_\lambda\left(\vpsi_{\lambda}^{\left(k\right)}, \phi^{\left(k\right)}\right) = \E_{\mathcal{B}}\left[\max\left(0, \lambda\left(\vs;\vpsi_{\lambda}^{\left(k\right)},\vphi^{\left(k\right)}\right)\right)\left(\hat{Q_{r}}\left(\vs,\va;\vpsi_{r}^{\left(k\right)},\vphi^{\left(k\right)}\right)-\target{V_{r}}\right)\right]$} \\ \Comment{\Eqref{eq:constraint_loss}, no gradient through $\hat{Q_{r}}$}
\State{$\vpsi_{r,c,\lambda}^{\left(k+1\right)}, \phi^{\left(k+1\right)} = \vpsi_{r,c,\lambda}^{\left(k\right)}, \phi^{\left(k\right)} - \eta_1 \cdot \nabla_{\vpsi_{r,c,\lambda}^{\left(k\right)}, \phi^{\left(k\right)}} \sum_{j \in \left\{r, c, \lambda\right\}} L_{j}\left(\vpsi_{j}^{\left(k\right)}, \phi^{\left(k\right)}\right)$}
\State{\bf Policy update:}
\State{$\vtheta^{\left(k+1\right)} = \vtheta^{\left(k\right)} + \eta_2 \cdot \E_{\mathcal{B}}\left[\text{policyGradient}\left(\vtheta^{\left(k\right)}, \vs,\va, Q_{\lambda}\left(\vs,\va;\vpsi_{r,c,\lambda}^{\left(k\right)},\vphi^{\left(k\right)}\right)\right)\right]$}
\Until{stopping criterion is met}
\State return $\vpsi_{r,c,\lambda}^{\left(k+1\right)}$, $\vphi^{\left(k+1\right)}$ and $\vtheta^{\left(k+1\right)}$
\end{algorithmic}
\end{algorithm}

\subsection{Policy Evaluation}

Our method needs to have access to a Q-function for optimization.
While any method for policy evaluation can be used, we rely on the Retrace algorithm \citep{munos2016safe}.
More concretely, we learn the Q-function for each cost term $Q_{i}\left(\vs,\va; \vpsi_{i},\vphi\right)$, where $\vpsi_{i},\vphi$ denote the parameters of the function approximator, by minimizing the mean squared loss:

\begin{equation}
\begin{aligned}
\min_{\vpsi_{i},\phi} L(\vpsi_{i}, \phi) &= \min_{\vpsi_{i},\vphi} \bE_{\mu_b(s), b(a|s)} \Big[ \big( Q_{i}\left(\vs_t,\va_t; \vpsi_{i},\vphi\right) - Q^{\text{ret}}_t \big)^2 \Big]\text{, with} \\ 
  Q^{\text{ret}}_t &= \!\begin{multlined}[t]
  Q_{i}\left(\vs_t,\va_t; \vpsi_{i}',\vphi'\right) +
  \sum_{j=t}^\infty \gamma^{j-t} \Big( \prod_{k=t+1}^j c_k \Big) \Big[ r_i(s_j, a_j) + \\
  \mathbb{E}_{\pi(a | s_{j+1})} [ Q_{i}\left(\vs_{j+1},\va; \vpsi_{i},\vphi\right) ] - Q_{i}\left(\vs_j,\va_j; \vpsi_{i}',\vphi'\right) \Big]\text{,}\end{multlined}\\
  c_k &= \min\Big(1, \frac{\pi(a_k | s_k)}{b(a_k | s_k)}\Big)\text{,}
\end{aligned}
\end{equation}

where $Q_{i}\left(\vs,\va; \vpsi_{i}',\vphi'\right)$ denotes the output of a target Q-network, with parameters $\vpsi_{i}',\vphi'$, that we copy from the current parameters after a fixed number of updates.
Note that while the above description uses the definition of reward $r_i$ we learn the value for the costs analogously.
We truncate the infinite sum after $N$ steps by bootstrapping with $Q_{\phi'}$.
Additionally, $b(a|s)$ denotes the probabilities of an arbitrary behaviour policy, in our case given through data stored in a replay buffer.

We use the same critic model to predict all values as well as the Lagrangian multipliers $\lambda\left(\vs,\vpsi_{\lambda},\vphi\right)$.
Following \Eqref{eq:constraint_loss}, we hence also minimize the following loss:

\begin{equation}
    \min_{\vpsi_{\lambda},\phi} L\left(\vpsi_{\lambda}, \phi\right) = \bE_{\mu_b(s)}\left[\min_{\lambda\left(\vs,\vpsi_{\lambda},\vphi\right) \geq 0} \E_{\rva \sim \pi}\left[Q_{\lambda}\left(\vs,\va\right)\right]\right]
\end{equation}

Our total critic loss to minimize is $\sum_{i} L\left(\vpsi_{i}, \phi\right) + \beta \cdot L\left(\vpsi_{\lambda}, \phi\right)$, where $\beta$ is used to balance the constraint and value prediction losses.

\subsection{Maximum a Posteriori Policy Optimization}

Given the Q-function, in each policy optimization step, MPO used  expectation-maximization (EM) to optimize the policy.
In the E-step MPO finds the solution to a following KL regularized RL objective; the KL regularization here helps avoiding premature convergence, we note, however, that our method would work with any other policy gradient algorithm for updating $\pi$.
MPO performs policy optimization via an EM-style procedure.
In the E-step a sample based optimal policy is found by minimizing: 

\begin{equation}
  \begin{aligned}
  & \max_q \bE_{\mu(s)} \Big[ \bE_{q(a|s)} \Big[  Q_{i}\left(\vs_t,\va_t; \vpsi_{i},\vphi\right) \Big] \Big] \\
  & s.t. \bE_{\mu(s)} \Big[ \textrm{KL}(q(a|s) , \pi_{old}(a|s) ) \Big] < \epsilon.
  \end{aligned}
\end{equation}

Afterwards the parametric policy is fitted via weighted maximum likelihood learning (subject to staying close to the old policy) given via the objective:

\begin{equation}
\begin{aligned}
\max_\pi \bE_{\mu(s)} \Big[ \bE_{q(a|s)} \Big[ \log \pi(a|s) \Big] \Big] \\
s.t.\ \bE_{\mu(s)} \Big[ \textrm{KL}(\pi_{old}(a|s) , \pi(a|s) ) \Big] < \epsilon\text{,}
\end{aligned}
\end{equation}

assuming a Gaussian policy (as in this paper) this objective can further be decoupled into mean and covariance parts for the policy (which in-turn allows for more fine-grained control over the policy change) yielding:

\begin{equation}
\begin{aligned}
\max_\pi \bE_{\mu(s)} \Big[ \bE_{q(a|s)} \Big[ \log \pi(a|s) \Big] \Big] \\
s.t.\ C_{\mu} < \epsilon_{\mu} \\
    \ C_{\Sigma} < \epsilon_{\Sigma}
\end{aligned}
\end{equation}

\begin{align}
\int \mu(s) \textrm{KL}(\pi_{old}(a|s) , \pi(a|s)) = C_\mu + C_\Sigma,
\end{align}
where
$$C_{\mu} = \int \mu(s) \tfrac{1}{2}(\textrm{tr}(\Sigma^{-1}\Sigma_{old}) - n + \ln(\frac{\Sigma}{\Sigma_{old}})) ds, $$
$$C_{\Sigma} = \int \mu(s) \tfrac{1}{2}(\mu -\mu_{old})^T\Sigma^{-1}(\mu -\mu_{old}) ds.$$

This decoupling of updating mean and covariance allows for setting different learning rate for mean and covariance matrix and controlling the contribution of the mean and co-variance to KL seperatly. 
For additional details regarding the rationale of this procedure we refer to the original paper \cite{abdolmaleki2018maximum}.

\newpage

\subsection{Hyperparameters}

The hyperparameters for the Q-learning and policy optimization procedure are listed in Table \ref{tab:hyperparameters}.
We perform optimization of the above given objectives via gradient descent; using different learning rates for critic and policy learning.
We use Adam for optimization.

\def\arraystretch{1.5}
\begin{table}[h!]
\caption{\small{Overview of the hyperparameters used for the experiments.}}
\label{tab:hyperparameters}
\centering
\begin{tabular}{c|c|c|c|c}
\multicolumn{1}{c}{\bf Parameter} &\multicolumn{1}{c}{\bf Cart-pole}  &\multicolumn{1}{c}{\bf Humanoid}  &\multicolumn{1}{c}{\bf Minitaur} &\multicolumn{1}{c}{\bf Sawyer}\\
\hline
Hidden units policy &$100-100$ &$300-200$ &$300-200$ &$200-200$\\
Hidden units critic &$200-200$ &$400-300$ &$300-200$ &$400-200$\\
LSTM cells &- &- &$100$ &-\\
Discount &$0.99$ &$0.99$ &$0.99$ &$0.99$\\
Policy learning rate &$\expnumber{1}{-5}$ &$\expnumber{1}{-5}$ &$\expnumber{1}{-5}$ &$\expnumber{5}{-4}$\\
Critic learning rate &$\expnumber{1}{-4}$ &$\expnumber{1}{-4}$ &$\expnumber{3}{-4}$ &$\expnumber{5}{-4}$\\
Constraint loss scale ($\beta$) &$\expnumber{1}{0}$  &$\expnumber{1}{0}$ &$\expnumber{1}{-3}$ &$\expnumber{1}{-4}$\\
Number of actors &$32$ &$32$ &$100$ &$1$\\
E-step constraint($\epsilon$) &$\expnumber{1}{-1}$ &$\expnumber{1}{-1}$ &$\expnumber{1}{-2}$ &$\expnumber{1}{-1}$\\
M-step constraint on $\mu$ ($\epsilon_\mu$) &$\expnumber{1}{-2}$ &$\expnumber{1}{-2}$ &$\expnumber{1}{-4}$ &$\expnumber{5}{-4}$\\
M-step constraint on $\Sigma$ ($\epsilon_\Sigma$) &$\expnumber{1}{-5}$  &$\expnumber{1}{-5}$ &$\expnumber{1}{-6}$ &$\expnumber{1}{-5}$\\
\end{tabular}
\end{table}

\section*{Appendix B: Minitaur simulation details}

\def\arraystretch{1.5}
\begin{table}[h!]
\caption{\small{Overview of the different model variations and noise models in the Minitaur domain.
$\normal{\mu}{\sigma}$ is the normal distribution, $\lognormal{\mu}{\sigma}$ the corresponding log-normal.
$\uniform{a}{b}$ is the uniform distribution and $\bernouilli{p}$ the Bernouilli distribution.}}
\label{tab:minitaur_variations}
\centering
\begin{tabularx}{\textwidth}{c|c|X}
\multicolumn{1}{c}{\bf Parameter} &\multicolumn{1}{c}{\bf Sample frequency} &\multicolumn{1}{c}{\bf Description}\\
\hline
Body mass &episode & global scale $\sim \lognormal{0}{0.1}$, with scale for each separate body $\sim \lognormal{0}{0.02}$\\
Joint damping &episode & global scale $\sim \lognormal{0}{0.1}$, with scale for each separate joint $\sim \lognormal{0}{0.02}$\\
Battery voltage &episode &global scale $\sim \lognormal{0}{0.1}$, with scale for each separate motor $\sim \lognormal{0}{0.02}$\\
IMU position &episode & offset $\sim \normal{0}{0.01}$, both cartesian and angular\\
Motor calibration &episode &offset $\sim \normal{0}{0.02}$\\
Gyro bias &episode &$\normal{0}{0.001}$\\
Accelerometer bias &episode &$\normal{0}{0.01}$\\
Terrain friction &episode &$\uniform{0.2}{0.8}$\\
Gravity &episode &scale $\sim \lognormal{0}{0.033}$\\
Motor position noise &time step &$\normal{0}{0.04}$, additional dropout $\sim \bernouilli{0.001}$\\
Angular position noise &time step &$\normal{0}{0.001}$\\
Gyro noise &time step &$\normal{0}{0.01}$\\
Accelerometer noise &time step &$\normal{0}{0.02}$\\
Perturbations &time step &Per-step decay of 5\%, with a chance $\sim \bernouilli{0.001}$ of adding a force $\sim \normal{0}{10}$ in any planar direction\\
\end{tabularx}
\end{table}

\end{document}